%% file: main.tex
\documentclass[10pt,twocolumn,letterpaper]{article}

\usepackage[pagenumbers]{cvpr} 

\input{preamble}
\definecolor{cvprblue}{rgb}{0.21,0.49,0.74}
\usepackage[pagebackref,breaklinks,colorlinks,allcolors=cvprblue]{hyperref}
\usepackage[utf8]{inputenc} 
\usepackage[T1]{fontenc}    
\usepackage{hyperref}       
\usepackage{url}            
\usepackage{booktabs}       
\usepackage{amsfonts}       
\usepackage{amsmath}        
\usepackage{nicefrac}       
\usepackage{microtype}      
\usepackage{xcolor}         
\usepackage{multirow}       
\usepackage{graphicx}       
\usepackage{xcolor}         
\usepackage{caption}        
\usepackage{array}          
\usepackage[most]{tcolorbox}
\usepackage{circledsteps}
\usepackage[table]{xcolor}
\usepackage{tabularx}               
\usepackage{arydshln} 
\usepackage{amssymb}
\usepackage{pifont}
\usepackage{algorithm}
\usepackage{algorithmic}
\usepackage{verbatim}
\usepackage{fancyhdr}

\definecolor{LightOrange}{rgb}{1.0, 0.85, 0.7}
\definecolor{lightblue}{HTML}{42bcf5}

\newcommand{\cmark}{\ding{51}}%
\newcommand{\xmark}{\ding{55}}%

\definecolor{sae_orange}{HTML}{D67454}
\definecolor{ntublue}{HTML}{181C62}

\tcbuselibrary{most}
\tcbset{
  aibox/.style={
    width=\textwidth,
    top=10pt,
    colback=white,
    colframe=black,
    colbacktitle=black,
    enhanced,
    center,
    attach boxed title to top left={yshift=-0.1in,xshift=0.15in},
    boxed title style={boxrule=0pt,colframe=white,},
  }
}
\newtcolorbox{AIbox}[2][]{aibox,title=#2,#1}

\usepackage{CJKutf8}
\usepackage[misc]{ifsym}

\definecolor{lowyellow}{RGB}{241, 196, 15}

\def\paperID{12226} 
\def\confName{CVPR}
\def\confYear{2026}

\title{OpenMMReasoner: Pushing the Frontiers for Multimodal Reasoning \\with an Open and General Recipe}

\author{Kaichen Zhang$^{*,1,2,4}$, Keming Wu$^{*,1,3,4}$, Zuhao Yang$^{1,2,4}$, Bo Li$^{2,4}$, Kairui Hu$^{2,4}$, \\
Bin Wang$^{3}$, Ziwei Liu$^{2,4}$, Xingxuan Li$^{1}$\textsuperscript{\Letter}, Lidong Bing$^{1}$\\
$^{1}$MiroMind AI, $^{2}$Nanyang Technological University, $^{3}$Tsinghua University,
$^{4}$LMMs-Lab Team \\
{Email Contact: \tt\small\{zhan0564\}@e.ntu.edu.sg, \tt\small\{xingxuan.li,lidong.bing\}@miromind.ai} \\
}

\definecolor{ntured}{HTML}{D71440}
\definecolor{ntublue}{HTML}{181C62}

\usepackage{CJKutf8}

\newif\ifshowcomments
\showcommentstrue      

\begin{document}

\twocolumn[{%
   \renewcommand\twocolumn[1][]{#1}%
   \maketitle
   \vspace{-30pt}
   \begin{center}
    \centering
    \centerline{\url{https://evolvinglmms-lab.github.io/OpenMMReasoner/}}
    \vspace{1em}
    \includegraphics[width=0.95\textwidth]{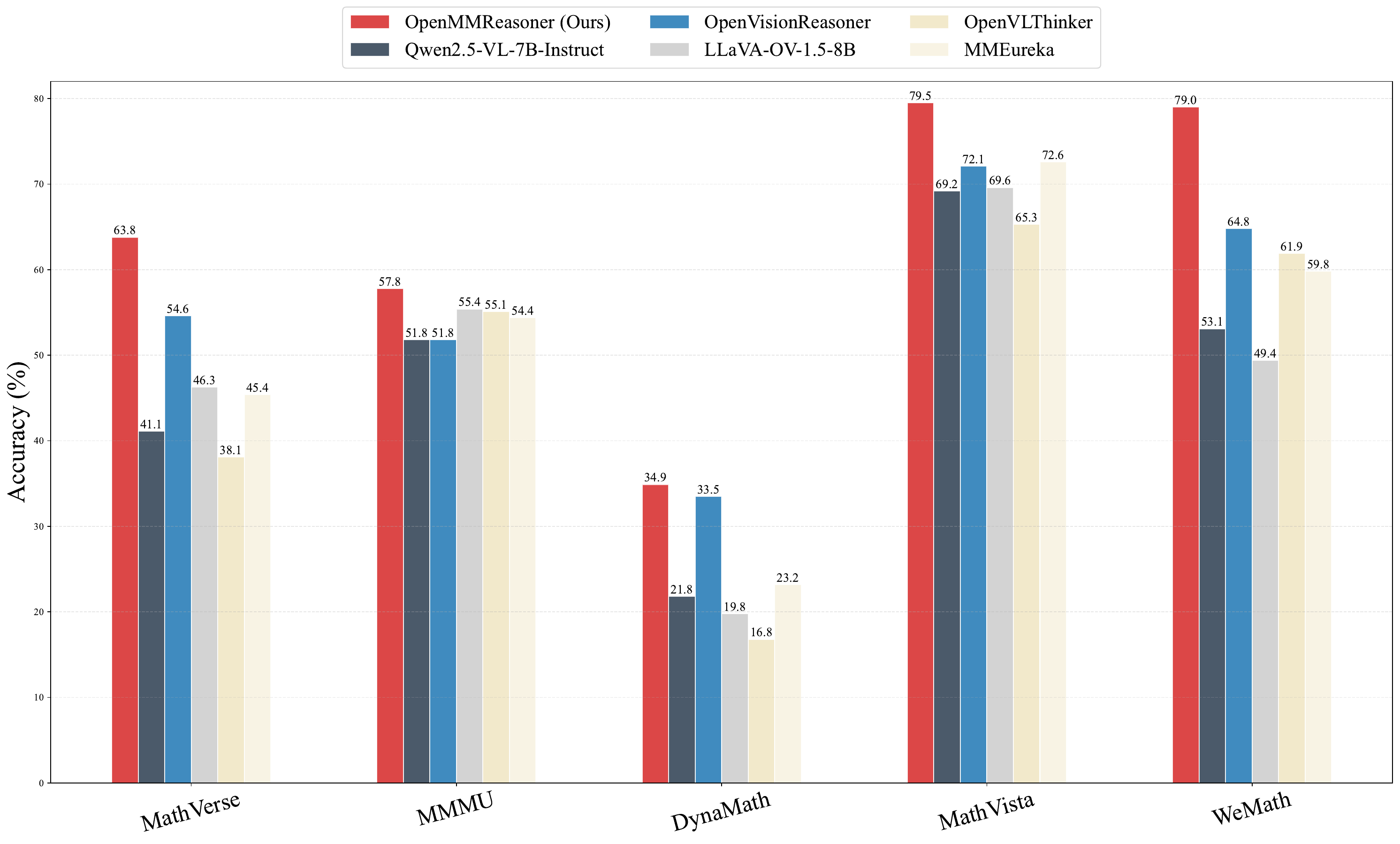}
    \vspace{-10pt}
    \captionof{figure}{\textbf{Performance Comparison with State-of-the-Art Large Multimodal Reasoning Models across Various Benchmarks.}
    Our proposed \textit{\textbf{OpenMMReasoner}} consistently outperforms competing methods, highlighting its effectiveness in complex reasoning tasks.}
    \label{fig:teaser}
   \end{center}%
}]

\fancypagestyle{firstpage}{
    \fancyhf{}
    \vspace{-3mm}
    \fancyhead[L]{%
        \vspace{2mm}
        \includegraphics[height=1cm]{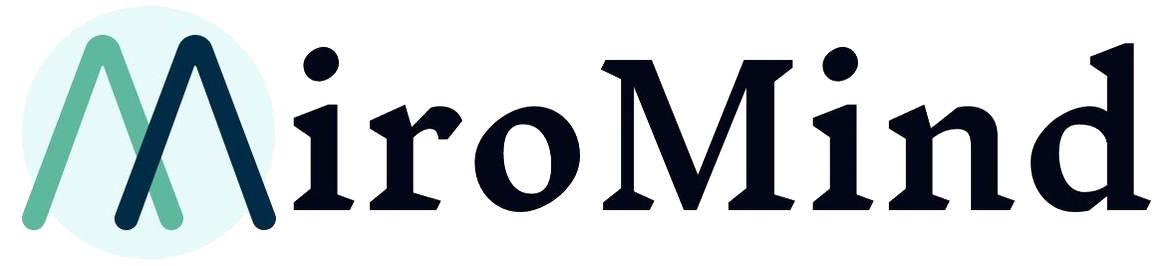}
        \vspace{-3mm}
    }
}
\thispagestyle{firstpage}

\renewcommand{\thefootnote}{\fnsymbol{footnote}}
\footnotetext[1]{Equal contribution. \textsuperscript{\Letter}Corresponding author.}
\renewcommand*{\thefootnote}{\arabic{footnote}}

\input{sec/0_abstract}    
\input{sec/1_intro}

\input{sec/2_related_work}

\input{sec/3_sft_recipe}

\input{sec/4_rl_recipe}

\input{sec/5_conclusion}

\clearpage
\newpage
{
    \small
    \bibliographystyle{ieeenat_fullname}
    \bibliography{main}
}

\input{sec/X_suppl}

\end{document}

%% file: sec/0_abstract.tex

\begin{abstract}
Recent advancements in large reasoning models have fueled growing interest in extending such capabilities to multimodal domains.
However, despite notable progress in visual reasoning, the lack of transparent and reproducible data curation and training strategies remains a major barrier to scalable research.
In this work, we introduce \textbf{OpenMMReasoner}, a fully transparent two-stage recipe for multimodal reasoning spanning supervised fine-tuning (SFT) and reinforcement learning (RL). In the SFT stage, we construct an 874K-sample cold-start dataset with rigorous step-by-step validation, providing a strong foundation for reasoning capabilities. The subsequent RL stage leverages a 74K-sample dataset across diverse domains to further sharpen and stabilize these abilities, resulting in a more robust and efficient learning process. Extensive evaluations demonstrate that our training recipe not only surpasses strong baselines but also highlights the critical role of data quality and training design in shaping multimodal reasoning performance. Notably, our method achieves a 11.6\% improvement over the Qwen2.5-VL-7B-Instruct baseline across nine multimodal reasoning benchmarks, establishing a solid empirical foundation for future large-scale multimodal reasoning research. We open-sourced all our codes, pipeline, and data at \url{https://github.com/EvolvingLMMs-Lab/OpenMMReasoner}. 

\end{abstract}
\vspace{-5mm}

%% file: sec/1_intro.tex
\section{Introduction}
\label{sec:intro}

With the rapid progress of reinforcement learning with verifiable rewards (RLVR), leading models such as DeepSeek-R1~\citep{dsr1_cite} and OpenAI o3~\citep{openai_o3_o4mini_2025} have demonstrated strong reasoning capabilities across mathematics, programming, and scientific tasks.
These advancements highlight RLVR as an effective paradigm for improving structured reasoning in large language models (LLMs).
Motivated by this progress, researchers have increasingly explored RL-enhanced reasoning from LLMs to large multimodal models (LMMs).
Recent studies~\citep{meng2025mm, chen2025sftrlearlyinvestigation, deng2025openvlthinkercomplexvisionlanguagereasoning, wei2025openvisionreasonertransferring} push the boundaries of multimodal reasoning, showing that reinforcement learning (RL) can strengthen fine-grained visual understanding and complex cross-modal problem-solving. These developments suggest that the benefits of RLVR can transfer beyond text, offering a scalable path toward more capable and reliable multimodal reasoning systems.

Despite these advancements, transparency across the training pipeline remains limited. While numerous studies investigate the supervised-finetuning (SFT) and RL stages~\citep{bai2025qwen25vltechnicalreport, openai_o3_o4mini_2025, comanici2025gemini}, few provide details of their data curation processes or conduct comprehensive ablation analyses. This lack of openness restricts reproducibility and obscures a deeper understanding of how reasoning-capable LMMs are actually built and how their training dynamics evolve. Recent efforts such as ThinkLite-VL~\citep{wang2025thinklite} and MM-Eureka~\citep{meng2025mm} propose clearer methodologies for RL data construction and offer useful insights for algorithm design. Other work, including Open Vision Reasoner (OVR)~\citep{wei2025openvisionreasonertransferring}, attempts to address both SFT and RL, but does not provide a scalable, unified recipe that generalizes across tasks and modalities. Similarly, LLM-focused efforts~\citep{guha2025openthoughtsdatarecipesreasoning} emphasize SFT while offering limited discussion of RL or multimodal extensions. Collectively, these limitations point to a critical gap: the lack of a transparent, scalable, and training recipe that unifies SFT and RL for building multimodal reasoning systems.

In this work, we present a comprehensive and scalable empirical study on training Large Multimodal Reasoning Models (LMRMs) from open-source LMMs (e.g., Qwen2.5-VL~\citep{bai2025qwen25vltechnicalreport}). Through extensive experimentation, we develop simple yet effective data pipelines for constructing high-quality SFT and RL datasets, enabling reliable cold-start training of a strong LMRM. Our analysis provides detailed insights into which components most effectively contribute at each stage, revealing practical principles for building robust, generalizable, and efficient multimodal reasoning systems. Furthermore, we explore how to maximize scaling efficiency across both SFT and RL phases and summarize key lessons learned throughout the process.



Building on these insights, we introduce \textbf{OpenMMReasoner}, a fully transparent training recipe for multimodal reasoning. It features (1) a high-quality 874k-sample SFT dataset that establishes a strong reasoning foundation prior to RL, and (2) a 74k-sample RL dataset accompanied by detailed analysis of diverse designs and optimization strategies, which further sharpens and stabilizes these capabilities. As shown in~\cref{fig:teaser}, our recipe yields a highly capable LMRM that consistently outperforms state-of-the-art methods such as OVR across a wide range of multimodal reasoning benchmarks, demonstrating both the effectiveness and scalability of our approach. The complete pipeline, spanning both the SFT and RL phases, is illustrated in~\cref{fig:data-pipeline}. As summarized in~\cref{tab:openness_comparison}, all components of our workflow are fully open and transparent, providing a comprehensive and reproducible view of the entire data curation and training process.

To summarize, our contributions are threefold:
\textbf{(1)} \textbf{\emph{Comprehensive Data Curation Insights.}}
We present the first systematic study on curating high-quality SFT and RL data for multimodal reasoning, supported by extensive and rigorous experiments across diverse modalities and reasoning types. We found that scaling data diversity is a critical factor for curating high-quality datasets. While diversity in data sources is important, diversity in answers represents an additional essential axis for improvement.
\textbf{(2)} \textbf{\emph{A Strong SFT Recipe for Reasoning.}}
We present a robust and reproducible SFT recipe that effectively equips LMMs with strong reasoning capabilities. Our approach incorporates step-by-step validation and offers practical insights for scaling a high-quality data pipeline focused on reasoning. By carefully selecting an appropriate teacher model for rejection sampling, and incorporating cross-domain data sources, we construct a cold-start dataset that exhibits both high diversity and quality, forming a solid reasoning foundation for subsequent RL.
\textbf{(3)} \textbf{\emph{An Advanced RL Recipe for Reasoning Enhancement.}}
We conduct a comprehensive comparative analysis of multiple RL strategies, including GSPO~\citep{zheng2025groupsequencepolicyoptimization}, GRPO~\citep{shao2024deepseekmathpushinglimitsmathematical}, and DAPO~\citep{yu2025dapoopensourcellmreinforcement}, to evaluate their stability, efficiency, and scaling behavior. By selecting the most suitable algorithm, we establish a robust RL pipeline that further sharpens and stabilizes reasoning abilities, delivering both high stability and superior performance.




\input{tab/compare_framework}

%% file: tab/compare_framework.tex
\begin{table}[t]
\centering
\resizebox{0.46\textwidth}{!}{%
\begin{tabular}{lcccc}
\toprule
\textbf{Method} & \textbf{Data Pipeline} & \textbf{SFT Data} & \textbf{RL Data} & \textbf{Model} \\
\midrule
\midrule
 Qwen2.5-VL-7B-Instruct~\citep{bai2025qwen25vltechnicalreport} & \xmark & \xmark & \xmark & \cmark  \\
 M2-Reasoning~\citep{an2025llava} &  \xmark & \xmark & \xmark & \cmark  \\
 MiMo-VL~\citep{coreteam2025mimovltechnicalreport} & \xmark & \xmark & \xmark & \cmark  \\
 OpenVisionReasoner~\citep{wei2025openvisionreasonertransferring} & \xmark & \cmark & \xmark & \cmark  \\
\midrule
\rowcolor{gray!10} \textbf{OpenMMReasoner (ours)}
  & \cmark & \cmark & \cmark & \cmark \\
\bottomrule
\end{tabular}%
}
\vspace{-1mm}
\caption{
\textbf{Comparison of Extent of Open-Sourcing across Existing LMRMs.}
OpenMMReasoner is the \emph{first} study to fully open-source its data pipeline, SFT/RL datasets, model weights, enabling fully transparent reproduction.
}
\label{tab:openness_comparison}
\end{table}

%% file: sec/2_related_work.tex
\begin{figure*}
    \centering
    \includegraphics[width=0.95\linewidth]{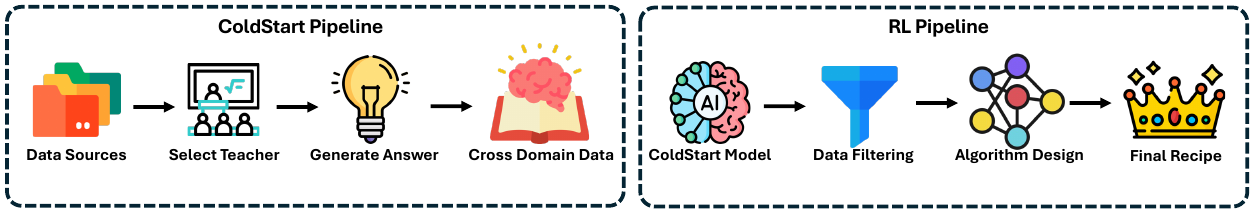}
    \caption{\textbf{Data Pipelines of OpenMMReasoner.} We propose two training recipes covering both the SFT and RL phases. The pipeline begins by collecting diverse data sources and selecting teacher models to generate new answer traces. During the RL phase, we explore different algorithm choices and filtering strategies, leading to our final optimized recipe.}
    \label{fig:data-pipeline}
    \vspace{-10pt}
\end{figure*}

\section{Related Work}
\label{sec:related_work}
RL has emerged as a powerful approach for enhancing reasoning in LLMs. Recent systems such as OpenAI’s o1~\citep{openaio1_cite} and DeepSeek-R1~\citep{dsr1_cite} have shown that multi-step reasoning and self-verification can emerge purely from large-scale RL. Building on these advances, recent works~\citep{meng2025mm,deng2025openvlthinkercomplexvisionlanguagereasoning,wei2025openvisionreasonertransferring,chen2025sftrlearlyinvestigation,leng2025mmr1enhancingmultimodalreasoning,qiao2025wemath20versatilemathbook,xu2025llavacotletvisionlanguage,yuan2025vlcogitoprogressivecurriculumreinforcement,yang2025longvt} extend RL-enhanced reasoning to multimodal models, while SFT studies~\citep{xu2025llavacotletvisionlanguage,yang2025r1onevisionadvancinggeneralizedmultimodal,li2025zebracotdatasetinterleavedvision,zhang2023videollamainstructiontunedaudiovisuallanguage,cheng2024videollama2advancingspatialtemporal,zhang2025videollama3frontiermultimodal} highlight the importance of high-quality supervision.

Despite this progress, many prior works~\citep{wei2025openvisionreasonertransferring,ai2025m2} lack transparent and reproducible training pipelines, and related efforts~\citep{guha2025openthoughtsdatarecipesreasoning} focus only on textual reasoning. This limits reproducibility and obscures how data curation and training design affect outcomes. To address this, we present OpenMMReasoner, a scalable and fully open recipe covering both SFT and RL, offering practitioners clear guidance for building reliable multimodal reasoning models.

%% file: sec/3_sft_recipe.tex
\vspace{-2mm}
\section{Supervised Fine-tuning Recipe}\label{sec:sft_recipe}

In this section, we present the data curation pipeline used to construct our SFT dataset. We begin with the initial data sources and describe how the dataset evolves through three stages: 1) data sourcing and formatting (103k raw questions), 2) data distillation and scaling (583k verified general-reasoning traces), and 3) cross-domain mixing to form the final 874k SFT recipe. The following experiments examine each stage and show how these design choices influence model performance. The distribution of the dataset across sources and domains is illustrated in~\cref{fig:data-distribution}.

\input{tab/data_source}

\input{tab/abl_teachers}
\input{tab/abl_data}

\subsection{Data Curation}

The overall pipeline is illustrated in~\cref{fig:data-pipeline}. The dataset is constructed through the following stages: \textbf{1)} Data Sourcing, \textbf{2)} Data Distillation and Scaling, and \textbf{3)} Domain Mixing.

\paragraph{Data Sourcing.}
We begin with approximately 103k raw question–answer pairs collected from public datasets including LLaVA-CoT~\citep{xu2025llavacotletvisionlanguage}, OpenVLThinker~\citep{deng2025openvlthinkercomplexvisionlanguagereasoning}, and We-Math2.0~\citep{qiao2024wemathdoeslargemultimodal}. These samples mainly cover general VQA and reasoning-oriented tasks. Additional sources used later in the domain mixing stage (Section~\ref{sec:domain_mixing}) include MMR1~\citep{zhang2024lmmsevalrealitycheckevaluation} for image-based mathematical reasoning and MiroMind-M1~\citep{li2025miromindm1opensourceadvancementmathematical} for text-based mathematical reasoning.

\paragraph{Data Formatting.}
Data sourced from different benchmarks often follow inconsistent answer styles and reasoning formats, which may introduce training instability~\citep{ye2025limoreasoning,li2025miromindm1opensourceadvancementmathematical}. We standardize all samples to a unified reasoning format, normalizing textual structure and ensuring consistent step-wise outputs before entering the distillation stage.

\subsection{Experimental Setup}

For the SFT training, we adopt online packing and the Liger-Kernel~\citep{hsu2025ligerkernel} to accelerate training efficiency. All experiments are conducted using the LMMs-Engine~\citep{lmms_engine2025} as the training framework and LMMs-Eval~\citep{lmms_eval2024,zhang2024lmmsevalrealitycheckevaluation} for standardized evaluation.
We use Qwen2.5-VL-7B-Instruct~\citep{bai2025qwen25vltechnicalreport} as our initial checkpoint to start, which also serves as our baseline. Each model is trained until convergence, and further implementation details are provided in the Appendix.

\input{tab/abl_filter}
\vspace{-1mm}

\subsection{Data Distillation}
\label{sec:teacher_model}


\paragraph{Teacher Model Selection.}
Prior work~\citep{dsr1_cite,guha2025openthoughtsdatarecipesreasoning} shows that distillation from a stronger model improves data quality and data efficiency. To identify the most suitable teacher, we compare answer traces distilled from multiple candidates, applying both format validation and answer verification.
Only samples whose final answers pass a rule-based validator and an LLM-as-judge check are retained, producing around 59k verified traces. As shown in \cref{tab:teacher}, all teacher-based variants outperform the baseline with a clear leading, confirming that a stronger teacher yields higher-quality supervision. Both of the stronger teacher models provide an average gain of at least 4.5 points across all benchmarks, with Qwen3-VL-235B-Instruct~\citep{qwen3vl2025} achieving the highest performance and selected as the teacher for distillation. This stage establishes the foundation for larger-scale answer sampling.

\paragraph{Scaling Data with verifiable answer.}
After selecting the optimal teacher, we enrich each question with multiple verified reasoning traces. We evaluate four scaling factors—$\times1$, $\times2$, $\times4$, and $\times8$—and observe consistent improvements as the number of verified answers increases (\cref{tab:sampling}). Quantitatively, increasing the number of verified answers per question from $\times1$ to $\times8$ raises the average benchmark performance from 50.5 to 55.2, demonstrating that answer diversity is a crucial factor for data quality and model generalization. We adopt the $\times8$ configuration as our final choice, as it provides substantial gains while keeping the computational cost of data generation manageable, representing a practical balance between performance and efficiency. The $\times8$ sampling configuration increase the dataset to roughly 583k verified general-reasoning samples. This expanded dataset serves as the core of our general SFT data before domain mixing.

\input{tab/abl_domain}
\subsection{Length and Difficulty Filtering}
\label{sec:data_filter}
We next examine whether further data refinement improves performance. Two common filtering strategies—difficulty-based filtering and length-based filtering—are evaluated. Difficulty is approximated from previous sampling accuracy, while token count is used to remove extremely short examples.
As shown in \cref{tab:filter}, both filtering strategies reduce performance, which we attribute to the loss of answer diversity. Since filtering decreases sample variety without improving consistency, we adopt a no-filtering policy for the final recipe, preserving the full 583k general-reasoning set produced from the distillation stage.

\input{tab/main_exp}
\subsection{Domain Mixing}
\label{sec:domain_mixing}
To further enhance reasoning generalization, we incorporate supervised data from additional reasoning domains. While the 583k distilled dataset already provides strong multimodal reasoning, its coverage of mathematical reasoning remains limited.
We therefore integrate MMR1~\citep{leng2025mmr1enhancingmultimodalreasoning} (image-based math reasoning) and MiroMind-M1~\citep{li2025miromindm1opensourceadvancementmathematical} (text-based math reasoning).
As shown in \cref{tab:domain-mix}, adding both types of mathematical supervision consistently improves performance across multimodal and reasoning benchmarks. Combining the general SFT data with these additional mathematical datasets yields our final 874k mixed SFT dataset.

\subsection{Analysis and Insights}
In summary, the dataset evolves through three major stages—from 103k raw questions, to 583k distilled general-reasoning samples, and finally to an 874k mixed SFT dataset incorporating both general and mathematical reasoning. As shown in~\cref{tab:main-exp}, our results across nine reasoning benchmarks~\citep{qiao2024wemathdoeslargemultimodal,qiao2024wemathdoeslargemultimodal,lu2024mathvistaevaluatingmathematicalreasoning,xiao2024logicvistamultimodalllmlogical,yue2024mmmumassivemultidisciplinemultimodal,yue2025mmmuprorobustmultidisciplinemultimodal,wang2024charxivchartinggapsrealistic,zou2025dynamathdynamicvisualbenchmark} show that our method achieves superior performance and data efficiency compared to other SFT approaches, demonstrating the effectiveness of our recipe in building a strong reasoning base model.

Our empirical findings highlight four key observations:
\textbf{(1) Answer diversity enhances reasoning.} Increasing the diversity of generated answers consistently improves the model’s overall reasoning performance, even when using the same question sources, suggesting that exposure to varied solutions strengthens understanding.  
\textbf{(2) Teacher model selection is crucial.} Distilling from a strong teacher model substantially boosts the model’s reasoning ability while maintaining high data efficiency. Careful selection for teacher model directly affects the quality of the distilled dataset and the final model performance.
\textbf{(3) Over-filtering reduces diversity and performance.} The best results are achieved without excessive filtering, indicating that maintaining greater answer diversity encourages more robust reasoning abilities.  
\textbf{(4) Cross-domain knowledge improves generalization.} Incorporating diverse data from multiple domains consistently enhances the model’s overall reasoning capabilities across tasks.

%% file: tab/data_source.tex
\definecolor{oursrow_orig}{HTML}{BFBFBA}
\colorlet{oursrow}{oursrow_orig!30}

\begin{figure}[t]
    \centering
    \includegraphics[width=0.98\linewidth]{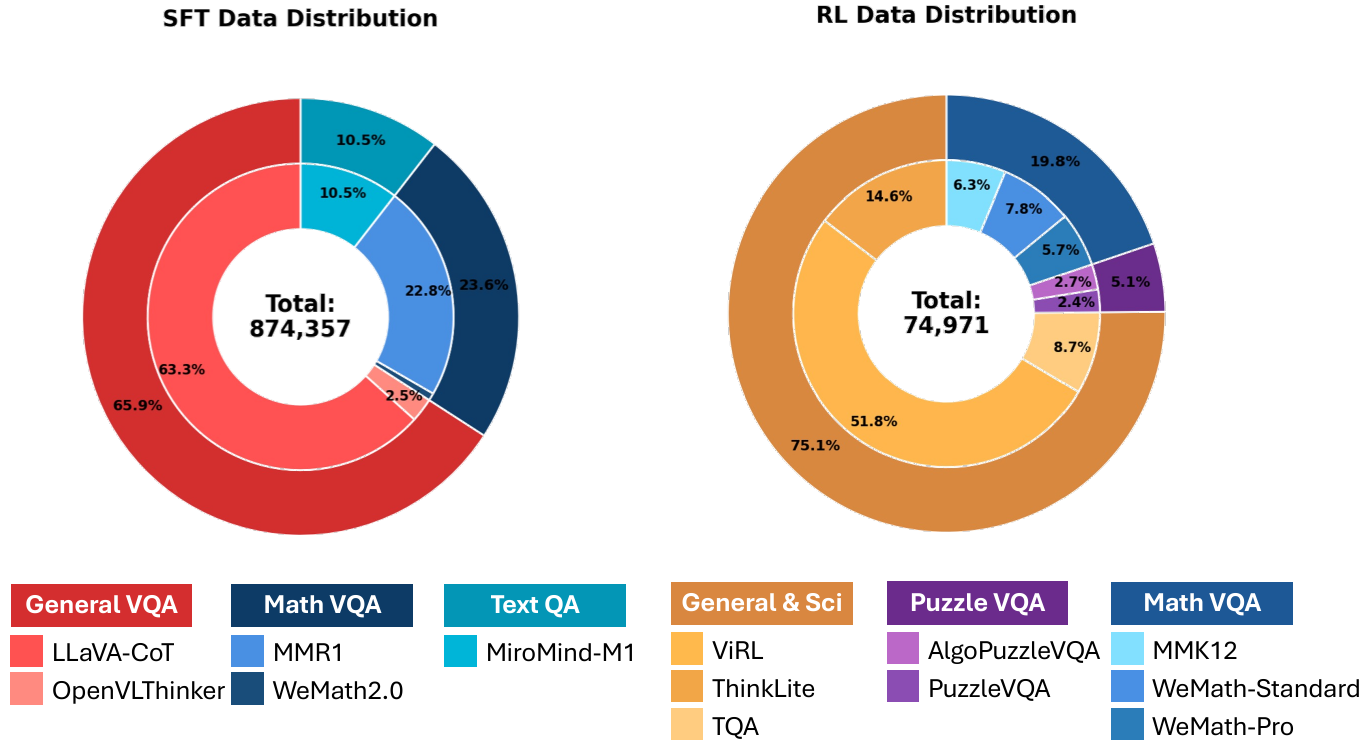}
    \vspace{-2mm}
    \caption{\textbf{Data Source Distribution OpenMMReasoner.} Our dataset comprises diverse sources across multiple domains, aiming to balance data diversity and efficiency for optimal performance. }
    \label{fig:data-distribution}
\end{figure}




%% file: tab/abl_teachers.tex
\begin{table}[t]
\resizebox{\linewidth}{!}{
\centering
\begin{tabular}{l  c  c c c}
\toprule
SFT Datasets & & \multicolumn{3}{c}{Benchmarks} \\
\midrule
\midrule
Teacher Model & Average & MathVision & MathVerse & MathVista \\
\midrule
\rowcolor{gray!20} Baseline & 45.3 & 25.5 & 41.1 & 69.2 \\
Qwen2.5-VL-72B-Instruct & 49.8 & 28.1 & 49.7 & \textbf{71.5} \\
Qwen3-VL-235B-Instruct & \textbf{50.5} & \textbf{29.3} & \textbf{52.9} & 69.4 \\
\bottomrule
\end{tabular}
}
\vspace{-5pt} 
\caption{\textbf{Teacher models improve performance and data efficiency.} Even with limited data, leveraging a teacher model significantly enhances the model’s reasoning ability at scale.}
\label{tab:teacher}
\end{table}

%% file: tab/abl_data.tex
\begin{table}
\centering
\resizebox{\linewidth}{!}{
\begin{tabular}{l  c  c c c}
\toprule
SFT Datasets & & \multicolumn{3}{c}{Benchmarks} \\
\midrule
\midrule
Sampling Strategy & Average & MathVision & MathVerse & MathVista \\
\midrule
$\times1$ sampling & 50.5 & 29.3 & 52.9 & 69.4 \\
$\times2$ sampling & 52.7 & 30.8 & 54.4 & 72.9 \\
$\times4$ sampling & 52.9 & 30.8 & 55.3 & 72.6 \\
$\times8$ sampling & \textbf{55.2} & \textbf{34.6} & \textbf{57.1} & \textbf{73.7} \\

\bottomrule
\end{tabular}
}
\vspace{-5pt}
\caption{\textbf{Repeated teacher sampling provides a scaling axis.} With the same question sources, increasing answer diversity improves data quality, showing that both question diversity and answer diversity are important for model performance.}
\label{tab:sampling}
\end{table}


%% file: tab/abl_filter.tex
\begin{table}
\centering
\resizebox{\linewidth}{!}{
\begin{tabular}{l  c  c c c}
\toprule
SFT Datasets & & \multicolumn{3}{c}{Benchmarks} \\
\midrule
\midrule
Filter & Average & MathVision & MathVerse & MathVista \\
\midrule
No Filter & \textbf{55.2} & \textbf{34.6} & \textbf{57.1} & \textbf{73.7}  \\ 
Length Filter & 54.2 & 33.0 & 56.0 & 73.6 \\
Difficulty Filter & 51.3 & 30.6 & 53.3 & 70.2  \\

\bottomrule
\end{tabular}
}
\vspace{-5pt}
\caption{\textbf{Over-filtering answers is detrimental.} Although filtering is commonly used in SFT, it reduces dataset diversity and does not improve overall performance.}
\label{tab:filter}
\end{table}

%% file: tab/abl_domain.tex
\begin{table}[t]
\centering
\resizebox{\linewidth}{!}{
\begin{tabular}{l  c  c c c}
\toprule
SFT Datasets & & \multicolumn{3}{c}{Benchmarks} \\
\midrule
\midrule
Domain-mixing & Average & MathVision & MathVerse & MathVista \\
\midrule
No Mix & 55.2 & 34.6 & 57.1 & 73.7  \\
ImgMath Mix & 55.6 & 34.3 & 57.5 & \textbf{74.9} \\
TxtMath Mix & 55.6 & 35.3 & 57.2 & 74.2 \\
Img+TxtMath & \textbf{56.3}  & \textbf{36.6} & \textbf{57.7}  & 74.8 \\

\bottomrule
\end{tabular}
}
\vspace{-5pt}
\caption{\textbf{Cross-domain data improves reasoning ability.} Mixing data from multiple domains enhances the model’s reasoning performance, demonstrating effective reasoning transfer across domains.}
\label{tab:domain-mix}
\end{table}
\vspace{-5pt}

%% file: tab/main_exp.tex
\begin{table*}[t]
\centering
\small
\caption{\textbf{Evaluation Results on Visual Reasoning Benchmarks.} 
Best results are \textbf{bold} and the second-best are \underline{underlined} for \textit{open-source models}. 
$^\dagger$ Indicates results reproduced by ourselves.}
\resizebox{\textwidth}{!}{%
\fontsize{22}{26}\selectfont
\begin{tabular}{lllccccccccccccc}
\toprule
\multirow{2}{*}{\textbf{Model}} & 
\textbf{SFT Data} &
\textbf{RL Data} &
\textbf{MathVista} & 
\multicolumn{2}{c}{\textbf{MathVision}} & 
\textbf{MathVerse} & 
\textbf{DynaMath} & 
\textbf{WeMath} & 
\textbf{LogicVista} & 
\multicolumn{1}{c}{\textbf{MMMU}} &
\multicolumn{2}{c}{\textbf{MMMU-Pro}} &
\multicolumn{2}{c}{\textbf{CharXiv}} & \\

& & & testmini & test & testmini & testmini & worst & loose & test & val & standard & vision & reas. & desc. \\

\midrule
\textbf{\textit{Close-source Models}}\\
OpenAI-GPT-4o~\cite{hurst2024gpt} & - & - & 59.9 & 31.1 & - & 40.6 & 34.5 & - & 64.4 & - & - & - & - & - \\
GPT-4o mini~\cite{hurst2024gpt} & - & - & 55.1 & 27.3 & - & 30.0 & 31.6 & 48.8 & 41.4 & - & 37.6 & - & 34.1 & 74.9 \\
\midrule
\textbf{\textit{SFT Methods}}\\

LLaVA-OneVision-7B~\cite{li2024llavaonevisioneasyvisualtask} & 4.8M & - & 62.6 & 17.6 & - & 17.6 & 9.0 & - & 32.0 & - & 24.1 & - & 23.6 & 48.7 \\
InternVL3-8B~\cite{zhu2025internvl3exploringadvancedtraining} & - & - & 70.5 & 28.6 & - & 33.9 & 23.0 & - & 43.6 & - & - & - & 37.6 & 73.6 \\
Qwen2.5-VL-7B~\cite{bai2025qwen25vltechnicalreport} & - & - & 69.2 & 25.5 & 25.6$^\dagger$ & 41.1 & 21.8 & 53.1 & 47.9 & 51.8$^\dagger$ & 37.9$^\dagger$ & 35.1$^\dagger$ & 36.4 & 67.3 \\
LLaVA-OneVision-1.5-8B~\cite{an2025llava} & 105M & -  & 69.6 & 25.6 & 21.7$^\dagger$ & 46.3 & 19.8$^\dagger$ & 49.4$^\dagger$ & 45.8 $^\dagger$ & 55.4 & 37.4 & 25.2 & 37.0$^\dagger$ & 74.1 \\
\rowcolor{LightOrange} \textbf{OMR-7B-ColdStart (ours)} & 874k & - & 74.8 & 36.6 & 33.9 & \underline{57.7} & 29.3 & 67.2 & 46.2 & 54.4 & 39.3 & 37.3 & 39.7 & \textbf{76.1} \\
\midrule

\textbf{\textit{RL-based Methods}}\\

VLAA-Thinker-Qwen2.5-7B~\cite{chen2025sftrlearlyinvestigation} & 126k & 25k & 68.0 & 26.4 & - & 48.2 & 22.4 & - & 48.5 & - & - & - & - & -  \\
ThinkLite-7B-VL~\cite{wang2025thinklite} & - & 11k & 71.6 & 24.6 & - & 42.9 & 16.5 & - & 42.7 & - & - & - & - & - \\
VL-Rethinker-7B~\cite{wang2025vlrethinkerincentivizingselfreflectionvisionlanguage} & - & 39k & 73.7 & 28.4 & - & 46.4 & 17.8 & - & 42.7 & - & 41.7 & - & - & - \\
M2-Reasoning~\cite{ai2025m2} & 6.2M & 102k & \underline{75.0} & 42.1 & - & 40.4 & - & - & \underline{50.6} & - & - & - & - & - \\
MMR1~\cite{leng2025mmr1enhancingmultimodalreasoning} & 1.6M & 15k & 72.0 & 31.8 & 29.0$^\dagger$ & 55.4 & 27.9$^\dagger$ & \underline{68.0}$^\dagger$ & 48.9 & 52.4$^\dagger$ & 41.1$^\dagger$ & 37.1$^\dagger$ & 43.5$^\dagger$ & 71.1$^\dagger$  \\
OpenVLThinker-7B~\cite{deng2025openvlthinkercomplexvisionlanguagereasoning} & 3.3k & 9.6k & 65.3 & 23.0 & 26.9$^\dagger$ & 38.1 & 16.8 & 61.9$^\dagger$ & 44.5 & \underline{55.1}$^\dagger$ & 39.7$^\dagger$ & \underline{38.4}$^\dagger$ & 41.0$^\dagger$ & 69.2$^\dagger$  \\
MM-Eureka-Qwen-7B~\cite{meng2025mm} & - & 15.6k & 72.6 & 28.1 & 32.1$^\dagger$ & 45.4 & 23.0 & 59.8$^\dagger$ & 46.3 & 54.4$^\dagger$ & 40.1$^\dagger$ & 37.1$^\dagger$ & 42.4$^\dagger$ & \underline{74.1}$^\dagger$  \\
OVR-7B~\cite{wei2025open} & 2M & 300k & 72.1 & \textbf{51.8} & \underline{38.2}$^\dagger$ & 54.6 & \underline{33.5} & 64.8 & \textbf{54.8} & 51.8$^\dagger$ & \textbf{50.2} & 29.1$^\dagger$ & \underline{44.5} & 73.6 \\
\rowcolor{LightOrange} \textbf{OMR-7B (ours)} & 874k & 74k & \textbf{79.5} & \underline{43.6} & \textbf{38.8} & \textbf{63.8} & \textbf{34.9} & \textbf{79.0} & 50.0 & \textbf{57.8} & \underline{44.1} & \textbf{40.6} & \textbf{46.1} & 73.5 \\
\bottomrule
\end{tabular}
}
\label{tab:main-exp}
\end{table*}

%% file: sec/4_rl_recipe.tex
\section{Reinforcement Learning Recipe}
\label{sec:rl_recipe}
To further enhance the model’s generalization and strengthen its multimodal reasoning capabilities, we introduce an RL phase tailored for multimodal reasoning tasks. Building on the strong reasoning foundation established during the SFT stage, this phase serves to further sharpen and stabilize these abilities. In this section, we provide key insights into our algorithmic design and the strategies employed to ensure stable and efficient RL dynamics.


\subsection{Preliminaries of Reinforcement Learning}
\label{sec:rl_pre}

Let each data pair $(q,a)$ be \textit{i.i.d} from a distribution $\mathcal{D}$, where $q$ is a query and $a$ is the ground-truth answer. 
Given an LLM policy $\pi_{\theta}(\cdot | \cdot)$, let $o$ be an LLM-generated response to $q$, and $r(\cdot,\cdot)$ is a predefined reward function that quantifies whether the response $o$ yields $a$. 
RL-based fine-tuning aims to maximize this expected reward over $\mathcal{D}$, \textit{i.e.,} 
\begin{align}
\max_{\theta}
    \mathcal{J}(\pi_{\theta})
    \triangleq
    \mathbb{E}_{(q,a)\sim \mathcal{D}}
    \mathbb{E}_{o\sim \pi_{\theta}(\cdot|q)}
    [r(o, a)].
    \label{eq:RL_obj}
\end{align}

\noindent{\textbf{Group Relative Policy Optimization (GRPO)}}\citep{shao2024deepseekmathpushinglimitsmathematical} is an efficient variant of PPO that removes the need for a critic network and Generalized Advantage Estimation (GAE), thereby reducing both memory usage and computational overhead. GRPO normalizes rewards within each rollout group to reduce variance and incorporates likelihood-ratio clipping with a KL-divergence penalty to constrain $\pi_\theta$ close to the initial SFT policy. The GRPO objective is defined as:

\vspace{-2mm}
\begin{small}
\begin{equation}
\label{eq:grpo}
\begin{aligned}
\mathcal{J}_\text{GRPO}(\theta) &=  \mathbb{E}_{(q,a)\sim\mathcal{D},\substack{\{o_i\}_{i=1}^G \sim \pi_{\theta_{\text{old}}} (\cdot \mid x)}} \\
&\biggl[ \frac{1}{G} \sum_{i=1}^G \frac{1}{|o_i|} \sum_{t=1}^{|o_i|} \biggl(\min \Bigl(r_{i,t}(\theta) \hat A_{i,t} \\
& \operatorname{clip}\left(r_{i,t}(\theta), 1-\varepsilon, 1+\varepsilon\right) \hat A_{i,t}\Bigl)  \biggr]
\end{aligned}
\end{equation}
\end{small}

\noindent{\textbf{Decoupled Clip and Dynamic Sampling Policy Optimization (DAPO)}}~\citep{yu2025dapoopensourcellmreinforcement} addresses several limitations of GRPO, including entropy collapse, training instability, and length bias caused by sample-level loss. To mitigate these issues, DAPO introduces a decoupled clipping mechanism and a dynamic sampling strategy, leading to a more stable and balanced optimization objective:

\vspace{-2mm}
\begin{small}
\begin{equation}
\label{eq:dapo}
\begin{aligned}
\mathcal{J}_\text{DAPO}(\theta) &=  \mathbb{E}_{(q,a)\sim\mathcal{D},\substack{\{o_i\}_{i=1}^G \sim \pi_{\theta_{\text{old}}} (\cdot \mid x)}} \\
&\biggl[ \frac{1}{\sum_{i=1}^{G}|o_i|} \sum_{i=1}^G \sum_{t=1}^{|o_i|} \biggl(\min \Bigl(r_{i,t}(\theta) \hat A_{i,t} \\
& \operatorname{clip}\left(r_{i,t}(\theta), 1-\varepsilon_{low}, 1+\varepsilon_{high}\right) \hat A_{i,t}\Bigl)  \biggr]
\end{aligned}
\end{equation}
\end{small}
\begin{small}
    subject to $0 < |\{o_j \mid \text{is\_equivalent}(a, o_i)\}| < G$
\end{small}

\noindent{\textbf{Group Sequence Policy Optimization (GSPO)}}~\citep{zheng2025groupsequencepolicyoptimization} tackles the token-level importance bias inherent in GRPO by introducing a sequence-level importance ratio for optimization. Additionally, GSPO employs a smaller clipping threshold $\varepsilon$ to enhance training stability. The resulting optimization objective for GSPO is defined as:

\begin{small}
\begin{equation}
\label{eq:gspo}
\begin{aligned}
\mathcal{J}_\text{GSPO}(\theta) &=  \mathbb{E}_{(q,a)\sim\mathcal{D},\substack{\{o_i\}_{i=1}^G \sim \pi_{\theta_{\text{old}}} (\cdot \mid x)}} \\
&\biggl[ \frac{1}{G} \sum_{i=1}^G\biggl(\min \Bigl(s_{i}(\theta) \hat A_{i,t} \\
& \operatorname{clip}\left(s_{i}(\theta), 1-\varepsilon, 1+\varepsilon\right) \hat A_{i,t}\Bigl)  \biggr]
\end{aligned}
\end{equation}
\end{small}
\begin{small}
    where $s_i$ is the importance ratio based on sequence likelihood.
\end{small}

\input{tab/rl_ablation}
\subsection{Dataset Curation}

\paragraph{Dataset Sourcing} Similar to the SFT stage, we begin by collecting diverse data samples across multiple domains. Our sources include MMEureka~\citep{meng2025mm}, ViRL~\citep{wang2025vlrethinkerincentivizingselfreflectionvisionlanguage}, TQA~\citep{DBLP:conf/cvpr/KembhaviSSCFH17}, We-Math~\citep{qiao2025wemath20versatilemathbook}, PuzzleVQA~\citep{chia2024puzzlevqadiagnosingmultimodalreasoning}, AlgoPuzzleVQA~\citep{ghosal2024algopuzzlevqa}, and ThinkLiteVL~\citep{wang2025thinklite}, covering domains such as science, mathematics, charts, and puzzles.

\vspace{-2mm}
\paragraph{Dataset Cleaning} To ensure answer validity, we extract and verify the final answers from each dataset. We then deduplicate by computing both image and text similarities to remove redundant questions. The cleaned datasets are merged to form the final dataset, resulting in approximately 74k samples for the RL stage. Overall data statistics are shown in~\cref{fig:data-distribution}. Unless otherwise specified, all RL experiments are conducted using this dataset.

\subsection{Experimental Setup}
For the RL training stage, we utilize verl~\citep{sheng2024hybridflow} and vllm~\citep{kwon2023efficient} to accelerate the training process. The evaluation protocol remains consistent with that of the SFT stage, employing LMMs-Eval~\citep{lmms_eval2024,zhang2024lmmsevalrealitycheckevaluation} to ensure a unified and standardized evaluation setup. By default, we use temperature 1.0 and the top-performing checkpoint from the SFT phase as the starting point for RL, unless stated otherwise. The checkpoint corresponds to the best-performing model reported in~\cref{tab:domain-mix}, where cross-domain mixing of image and text-based math data were applied, along with $\times8$ sampling for answer traces.

\begin{figure*}
    \centering
    \includegraphics[width=0.95\linewidth]{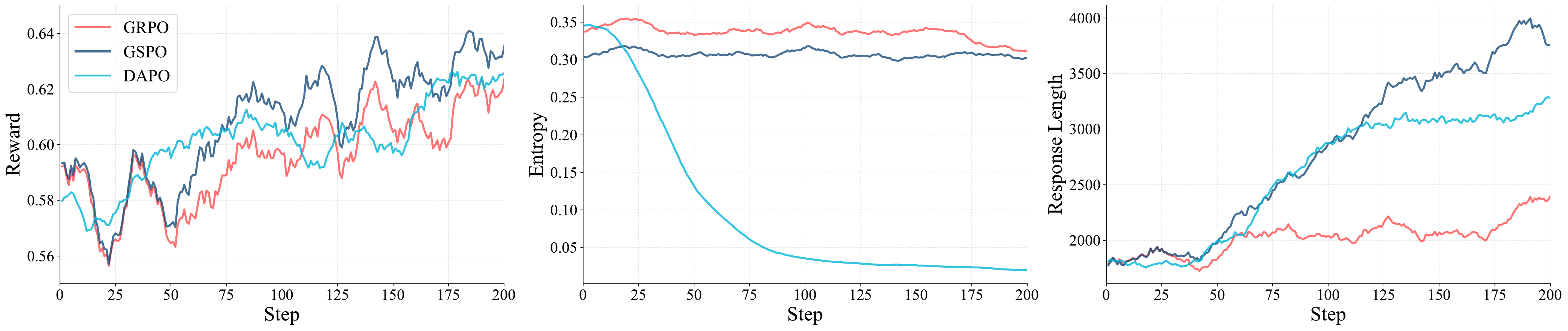}
    \vspace{-2mm}
    \caption{\textbf{Overall results across different algorithms.} We conduct a systematic comparison of various algorithms under identical multimodal RL training settings. GSPO demonstrates the highest training stability, exploration capability, and overall efficiency.}
    \label{fig:rl_curves}
\end{figure*}

\begin{figure*}[ht]
    \centering
    \includegraphics[width=0.98\linewidth]{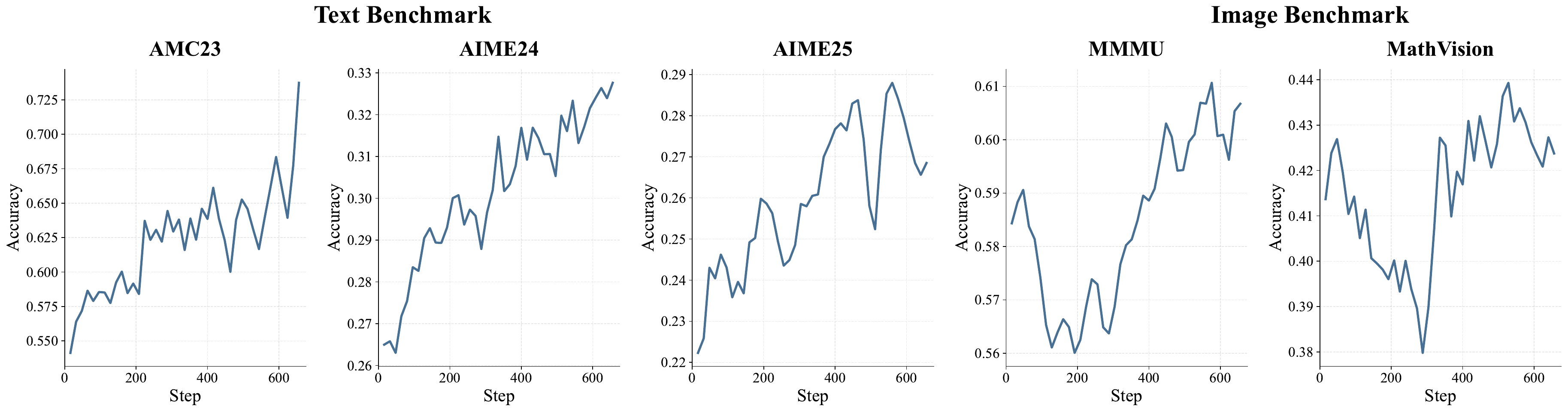}
    \vspace{-2mm}
    \caption{\textbf{Training dynamics on the validation set during RL.} During RL training, we observe that textual reasoning ability improves alongside visual reasoning, even when trained solely on multimodal data, indicating strong cross-domain generalization of reasoning capabilities.}
    \label{fig:val_curve}
\end{figure*}

\subsection{Training Recipe}
\label{sec:algo_selection}

\paragraph{RL Algorithm Selection}
To identify the most suitable RL algorithm for multimodal reasoning, we compare GSPO~\citep{zheng2025groupsequencepolicyoptimization}, DAPO~\citep{yu2025dapoopensourcellmreinforcement}, and GRPO~\citep{shao2024deepseekmathpushinglimitsmathematical} under a unified setting. We evaluate their training dynamics in terms of stability, exploration, and efficiency, as illustrated in~\cref{fig:rl_curves}. GSPO demonstrates faster convergence, higher rewards, and more stable behavior than DAPO and GRPO, achieving an effective balance between exploration and stability. While DAPO applies online filtering to remove zero-variance samples, it exhibits early entropy collapse and slower progress due to larger rollout requirements. GRPO shows moderate stability but converges more slowly. Based on these observations and the ablation results summarized in~\cref{tab:rl-abl}, we adopt GSPO as the RL algorithm in our final training setup.

\paragraph{Reward Function} To balance task accuracy and output formatting, we adopt a composite reward function similar to those in~\citep{dsr1_cite,openr1}, which combines both aspects in a weighted manner. Specifically, the final reward $R$ for each sample is defined as:
\begin{equation}
R = (1 - \lambda_{\text{fmt}}) \cdot R_{\text{acc}} + \lambda_{\text{fmt}} \cdot R_{\text{fmt}},
\end{equation}
where $R_{\text{acc}}$ measures the model's correctness on the task objective, and $R_{\text{fmt}}$ captures the consistency of the answer format. The coefficient $\lambda_{\text{fmt}} \in [0, 1]$ controls the trade-off between task accuracy and format adherence.  We use $\lambda_{\text{fmt}}=0.1$ throughout our RL experiments.

\vspace{0.5mm}

\begin{figure}
    \centering
    \includegraphics[width=0.95\linewidth]{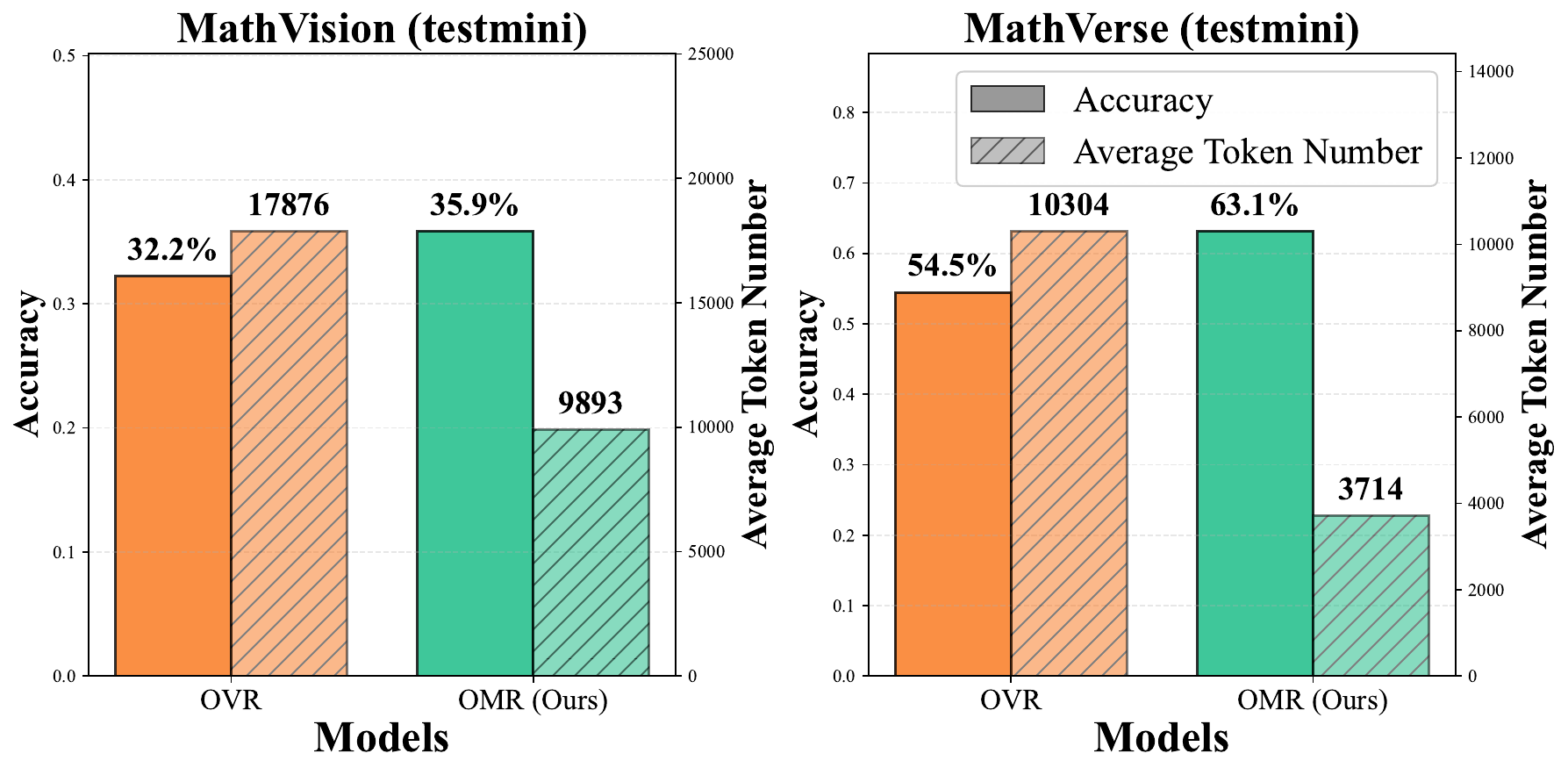}
    \caption{\textbf{Token efficiency comparison with OVR.} OpenMMReasoner achieve better accuracy while using significantly less token budget.}
    \label{fig:efficiency}
\end{figure}

\noindent\textbf{Sampling Strategies} Although RL algorithms naturally produce stronger learning signals on more challenging problems, their training efficiency remains limited. Previous work~\citep{kimiteam2025kimik15scalingreinforcement} employs a difficulty-based sampler, starting with easy tasks and gradually shifting to harder ones. We first use Qwen3-VL-8B~\citep{qwen3vl2025} to track the pass rate of each problem as a difficulty metric and implement the following strategy: RL training begins with simpler tasks and progressively transitions to more complex ones. Ablation results in~\cref{tab:rl-abl} show that curriculum learning does not outperform the original mixed sampling approach; therefore, we adopt the no-sampling strategy in our final configuration.

\paragraph{Balancing Reasoning Capability and Efficiency.} Our empirical findings show that while OpenVisionReasoner~\citep{wei2025openvisionreasonertransferring} achieves competitive performance, its response length becomes excessively long, raising concerns about reasoning efficiency. This observation highlights an important question: how can a model maintain strong reasoning performance while remaining computationally efficient and adaptable across different problem types.
In our reinforcement learning implementation, we adopt a similar overlength penalty strategy as proposed in DAPO~\citep{yu2025dapoopensourcellmreinforcement} to mitigate the overthinking behavior and achieve a balanced trade-off between reasoning depth and efficiency. As shown in \cref{fig:efficiency}, we evaluate models on two benchmarks, MMMU~\citep{yue2024mmmumassivemultidisciplinemultimodal} and We-Math~\citep{qiao2024wemathdoeslargemultimodal}. Our model demonstrates higher reasoning efficiency compared to OVR~\citep{wei2025openvisionreasonertransferring}. As shown in~\cref{fig:efficiency}, OVR requires excessively long reasoning trajectories to reach correct answers, whereas our model achieves a favorable balance between accuracy and efficiency, maintaining a reasonable reasoning budget while delivering superior overall performance.

\subsection{Analysis and Insights}
\label{sec:rl_analysis}

\paragraph{Overall Results.} We present the evaluation results of our final models for both the SFT and RL stages in~\cref{tab:main-exp}. Building on the strong reasoning foundation established during the SFT stage, the RL phase further sharpens and stabilizes these capabilities, leading to improved and more consistent performance. After RL, our model achieves state-of-the-art results on benchmarks such as WeMath~\citep{qiao2024wemathdoeslargemultimodal}, MathVerse~\citep{zhang2024mathversedoesmultimodalllm}, and MathVista~\citep{lu2024mathvistaevaluatingmathematicalreasoning} and demonstrating consistent improvement compare to SFT.
\paragraph{Textual Reasoning Transfers Alongside Strengthened Multimodal Reasoning.} 
As the model’s overall reasoning ability improves through RL training, we observe the gradual emergence of textual reasoning behaviors, suggesting a transfer of reasoning competence from multimodal to purely linguistic domains. As shown in~\cref{fig:val_curve}, validation performance on AIME24, AIME25, and AMC23 steadily increases throughout training, reflecting continuous and measurable gains in text-based reasoning capabilities. For AIME24 and AIME25, the reported accuracy represents the average over eight rollout runs. Prior work~\citep{huan2025doesmathreasoningimprove} has demonstrated that enhancements in mathematical reasoning can positively transfer to other reasoning domains. Evaluation results in~\cref{tab:text-emerge} compare baseline and RL-trained models, showing that textual reasoning improves and strengthens in tandem with multimodal reasoning across all training stages.  
These findings extend previous observations, indicating that cross-domain reasoning skills acquired via multimodal RL can effectively transfer to purely textual tasks, further highlighting the shared cognitive foundations and underlying mechanisms across different modalities.

\input{tab/abl_text_emerge}

\begin{figure}[t]
    \centering
    \includegraphics[width=0.9\linewidth]{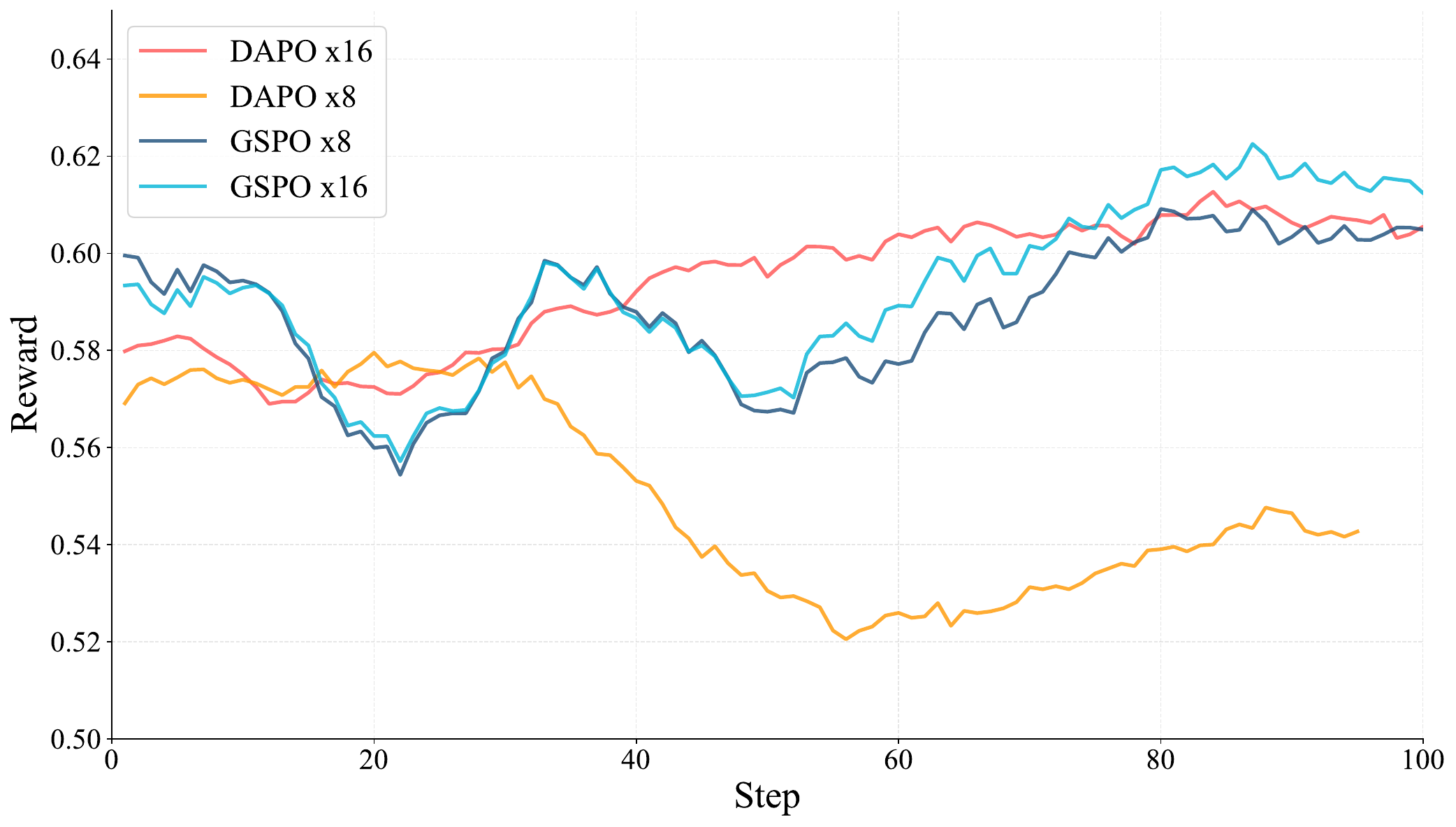}
    \vspace{-2mm}
    \caption{\textbf{Effect of rollout count on multimodal RL training stability.} DAPO becomes unstable with fewer rollouts, while increasing the rollout count leads to more stable training dynamics.}
    \label{fig:rollout}
\end{figure}

\vspace{-3mm}
\paragraph{Key Factors for Stable Training Dynamics.}
We identify two factors that critically affect RL training stability: rollout temperature and rollout count. First, higher rollout temperatures (e.g., 1.4) cause significant instability and occasional divergence, suggesting that excessive exploration amplifies policy gradient variance and destabilizes optimization. Second, the number of rollouts per update is central to maintaining convergence stability. To assess its impact, we compare $\times 8$ and $\times 16$ rollout configurations under both GRPO and DAPO. As shown in~\cref{fig:rollout}, the $\times 16$ configuration consistently yields higher rewards and smoother dynamics. This effect is especially pronounced for DAPO, where the $\times 8$ setting exhibits severe late-stage instability: entropy initially fluctuates and then abruptly spikes, ultimately causing training collapse. Interestingly, although larger rollout counts seem more computationally expensive, actual wall-clock time remains comparable between $\times 8$ and $\times 16$ due to similar token-length constraints. Based on these observations, we adopt a moderate temperature and a $\times 16$ rollout configuration as our default setup.

In summary, our key observations from the RL phase are as follows:
\textbf{1) GSPO outperforms other algorithms.} GSPO demonstrates superior stability and faster convergence compared to alternative methods in multimodal RL training.  
\textbf{2) Token efficiency is crucial.} While increasing reasoning steps at test time can improve performance, excessive tokens reduce efficiency. Our results show that a smaller reasoning budget can achieve comparable or even better accuracy.  
\textbf{3) Reasoning ability transfers across domains.} Gains in reasoning during training consistently translate into stronger performance across multiple domains.

%% file: tab/rl_ablation.tex
\begin{table*}[t]
\centering
\small
\caption{\textbf{Recipe selection results of different RL training strategies and coldstart starting point on Visual Reasoning Benchmarks.} Unless otherwise specified, the rollout temperature is set to 1.0 by default.} 
\vspace{-3mm}
\resizebox{\textwidth}{!}{%
\fontsize{22}{26}\selectfont
\begin{tabular}{lcccccccccccccc}
\toprule
\multirow{2}{*}{\textbf{Method}} & 

\textbf{Avg} & \textbf{MMMU}   & \multicolumn{2}{l}{\textbf{MMMU\_Pro}} & \textbf{MathVista} & \textbf{MathVerse} & \multicolumn{2}{l}{\textbf{MathVision}} & \multicolumn{2}{l}{\textbf{CharXiv}} & \textbf{LogicVista} & \textbf{WeMath} & \textbf{DynaMath}\\

&             & val             & standard            & vision           & testmini           & test               & testmini             & test             & reas              & desc             &              test       & loose           & worst             \\

\midrule

\rowcolor{gray!20}\multicolumn{14}{l}{\textbf{\textit{Coldstart Start Point:}} $\times8$ sampling + ImgTxt Math} \\
Baseline & 49.4         & 54.4            & 39.3                & 37.3             & 74.8               & 57.7               & 33.9                 & 36.6             & 39.7              & 76.1             & 46.2                & 67.2            & 29.3              \\
DAPO + $\times8$ rollout & 45.6 & 53.3 & 37.7 & 34.6 & 71.7 & 51.0 & 28.6 & 27.9 & 40.7 & 73.0 & 44.4 & 58.3 & 26.4 \\
DAPO + $\times16$ rollout & 48.9 & 52.9 & 40.6 & 35.6 & 74.3 & 56.2 & 30.3 & 34.5 & 43.4 & 77.0 & 46.9 & 67.6 & 27.4 \\
GRPO + $\times16$ rollout & 51.1 & 54.6 & 42.8 & 39.4 & 77.1 & 58.3 & 33.9 & 37.1 & 43.1 & 73.8 & 51.1 & 70.2 & 32.1 \\
GSPO + $\times8$ rollout & 51.6 & 54.9 & 41.2 & 38.8 & 76.9 & 61.4 & 35.9 & 39.9 & 45.0 & 73.8 & 46.9 & 71.7 & 32.3 \\
GSPO + $\times16$ rollout & 54.3 & 57.8 & 44.1 & 40.6 & 79.5 & 63.8 & 38.8 & 43.6 & 46.1 & 73.5 & 50.0 & 79.0 & 34.9 \\
GSPO + $\times16$ rollout + curriculum & 52.5 & 58.7 & 42.8 & 42.4 & 76.3 & 61.7 & 35.2 & 41.4 & 45.5 & 72.9 & 49.8 & 75.4 & 28.3 \\
\midrule
\rowcolor{gray!20}\multicolumn{14}{l}{\textbf{\textit{Coldstart Start Point:}}$\times8$ sampling} \\
Baseline & 49.2         & 55.6            & 40.6                & 37.7             & 73.7               & 57.1               & 35.2                 & 34.6             & 39.5              & 74.1             & 48.9                & 69.1            & 24.0              \\
DAPO + $\times16$ rollout & 49.6         & 54.8            & 42.0                & 37.6             & 74.1               & 57.8               & 36.8                 & 33.9             & 39.1              & 74.3             & 48.7                & 71.6            & 24.0              \\
DAPO + $\times16$ rollout + temp. 1.4 & 49.3 & 56.9 & 41.0 & 37.3 & 73.3 & 57.8 & 33.2 & 32.2 & 40.1 & 75.0 & 51.6 & 70.2 & 23.4 \\
GSPO + $\times16$ rollout & 51.0         & 55.1            & 42.4                & 39.6             & 75.1               & 59.4               & 36.2                 & 35.4             & 44.6              & 73.0             & 49.1                & 74.0            & 27.9    \\
GSPO + $\times16$ rollout + temp. 1.4 & 7.4  & 13.4 & 3.9  & 28.3 & 3.4  & 22.0 & 0.0  & 0.0  & 0.2  & 0.0  & 8.3  & 0.2  & 9.2  \\
\midrule
\rowcolor{gray!20}\multicolumn{14}{l}{\textbf{\textit{Coldstart Start Point:}}$\times1$ sampling} \\
Baseline & 46.7         & 54.2            & 42.6                & 36.0             & 71.5               & 49.7               & 29.9                 & 28.2             & 39.0              & 73.1             & 47.8                & 63.9            & 25.0              \\
GRPO + $\times8$ rollout & 47.6         & 55.6            & 41.5                & 36.7             & 73.5               & 53.0               & 29.6                 & 28.1             & 42.3              & 72.0             & 44.9                & 69.1            & 24.6              \\
DAPO + $\times8$ rollout & 49.2         & 54.4            & 50.8                & 38.4             & 74.5               & 55.7               & 27.6                 & 28.3             & 43.2              & 72.0             & 48.7                & 69.0            & 27.9              \\
DAPO + $\times8$ rollout + curriculum & 47.0         & 56.0            & 40.5                & 36.8             & 71.6               & 51.2               & 32.9                 & 28.1             & 36.2              & 73.6             & 46.2                & 66.8            & 24.6              \\

\bottomrule
\end{tabular}
}
\label{tab:rl-abl}
\end{table*}

%% file: tab/abl_text_emerge.tex
\begin{table}[!ht]
\centering
\resizebox{\linewidth}{!}{
\begin{tabular}{l  c  c c c}
\toprule
 & & \multicolumn{3}{c}{Benchmarks} \\
\midrule
\midrule
Model & Average & AIME24 & AIME25 & GPQA Diamond \\
\midrule
Baseline & 15.1 & 6.7 & 6.7 & 31.8 \\ 
\rowcolor{lightblue!20} + ColdStart & 22.2 & 16.7 & 14.6 & 35.4 \\
\rowcolor{lightblue!20} + RL & \textbf{29.4} & \textbf{27.1} & \textbf{22.1} & \textbf{38.9} \\

\bottomrule
\end{tabular}
}
\vspace{-5pt}
\caption{\textbf{Text reasoning ability transfer alongside visual reasoning improvement.} As the model’s overall reasoning capability increases, its text-based reasoning ability also improves. Across different training stages, we observe significant gains in textual reasoning, indicating strong cross-domain generalization.}
\label{tab:text-emerge}
\end{table}

%% file: sec/5_conclusion.tex
\section{Conclusion}
\label{sec:conclusion}

In this work, we present \textbf{OpenMMReasoner}, a transparent, scalable framework for training LMRMs via unified SFT and RL stages. Our study systematically examines how data curation, sampling strategies, and RL design shape multimodal reasoning. Experiments show that (1) well-designed SFT data, even in limited quantity, builds a strong reasoning foundation, and (2) carefully curated RL datasets, combined with effective algorithms like GSPO, enhance reasoning stability and performance. Scaling data \textit{diversity} across domains and reasoning traces proves more valuable than merely increasing size. Structured sampling and difficulty-aware curricula improve efficiency, while well-defined reward signals strengthen reasoning precision and multimodal consistency. These insights highlight the importance of transparent, reproducible pipelines, and we hope \textbf{OpenMMReasoner} serves as a solid empirical foundation and open-source reference for scalable multimodal reasoning research.

%% file: sec/X_suppl.tex
\clearpage
\setcounter{page}{1}
\setcounter{section}{0}
\setcounter{table}{0}
\setcounter{figure}{0}
\maketitlesupplementary

\begin{figure*}[h]
    \centering
    \includegraphics[width=1\linewidth]{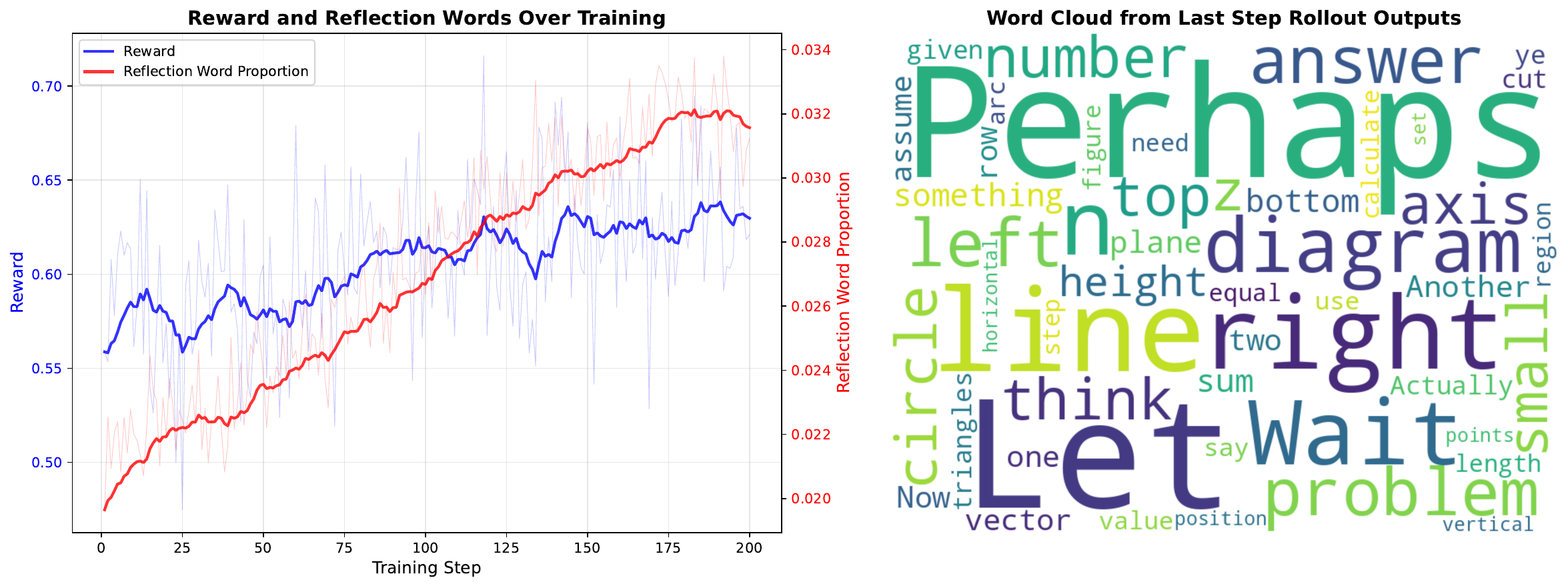}
    \caption{\textbf{Rollout Analysis over RL.} With the progress of RL training, the model response contains more reflection word ratio.}
    \label{fig:rollout_analysis}
\end{figure*}

\begin{table}[h]
\centering
\resizebox{0.38\textwidth}{!}{%
\begin{tabular}{lll}
\toprule
\textbf{Training Component} & \textbf{SFT}    & \textbf{RL}       \\
\midrule
optimizer          & AdamW   & AdamW     \\
scheduler          & cosine & constant \\
learning rate      & 5e-5   & 1e-6     \\
weight decay       & 0.0    & 0.1      \\
Training Steps     & 4300   & 1232      \\
Warmup Steps       & 430    & 25       \\
Max Length         & 61440  & 32792    \\
Dynamic Bsz        & True   & True     \\
Remove padding     & True   & True     \\
Liger Kernel       & True   & False   \\
\bottomrule
\end{tabular}
}
\caption{Detail parameters for SFT and RL training.}
\label{tab:train-detail}
\end{table}

\section{Implementation Details}

\begin{table*}[h]
\centering
\small
\caption{\textbf{Additional evaluation results on Sampling Strategy.} Scaling up with more answer generation leads to better result.}
\resizebox{\textwidth}{!}{%
\fontsize{22}{26}\selectfont
\begin{tabular}{lcccccccccccccc}
\toprule
\multirow{2}{*}{\textbf{Method}} & 

\textbf{Avg} & \textbf{MMMU}   & \multicolumn{2}{l}{\textbf{MMMU\_Pro}} & \textbf{MathVista} & \textbf{MathVerse} & \multicolumn{2}{l}{\textbf{MathVision}} & \multicolumn{2}{l}{\textbf{CharXiv}} & \textbf{LogicVista} & \textbf{WeMath} & \textbf{DynaMath}\\

&             & val             & standard            & vision           & testmini           & test               & testmini             & test             & reas              & desc             &              test       & loose           & worst             \\

\midrule
\rowcolor{gray!20}\multicolumn{14}{l}{\textbf{\textit{Sampling Strategy}}} \\
$\times1$ sampling & 46.7 & 54.2 & 42.6 & 36.0 & 71.5 & 49.7 & 29.9 & 28.2 & 39.0 & 73.1 & 47.8 & 63.9 & 25.0 \\
$\times2$ sampling & 47.5 & 54.4 & 39.1 & 34.6 & 72.9 & 54.4 & 30.9 & 30.8 & 37.5 & 73.6 & 49.6 & 69.3 & 23.0 \\
$\times4$ sampling & 48.2 & 56.9 & 41.0 & 37.6 & 72.6 & 55.3 & 30.3 & 30.8 & 40.7 & 75.4 & 45.8 & 69.3 & 23.0 \\
$\times8$ sampling & 49.2 & 55.6 & 40.6 & 37.7 & 73.7 & 57.1 & 35.2 & 34.6 & 39.5 & 74.1 & 48.9 & 69.1 & 24.0 \\

\bottomrule
\end{tabular}
}
\label{tab:extra_exp}
\end{table*}

\begin{table*}[h]
\centering
\small
\caption{\textbf{Reward ablation result.}}
\resizebox{\textwidth}{!}{%
\fontsize{22}{26}\selectfont
\begin{tabular}{lcccccccccccccc}
\toprule
\multirow{2}{*}{\textbf{Method}} & 

\textbf{Avg} & \textbf{MMMU}   & \multicolumn{2}{l}{\textbf{MMMU\_Pro}} & \textbf{MathVista} & \textbf{MathVerse} & \multicolumn{2}{l}{\textbf{MathVision}} & \multicolumn{2}{l}{\textbf{CharXiv}} & \textbf{LogicVista} & \textbf{WeMath} & \textbf{DynaMath}\\

&             & val             & standard            & vision           & testmini           & test               & testmini             & test             & reas              & desc             &              test       & loose           & worst             \\

\midrule
\rowcolor{gray!20}\multicolumn{14}{l}{\textbf{\textit{Reward $\lambda_{fmt}$ Settings}}} \\
$\lambda_{fmt} = 0.1$ & 51.1 & 54.6 & 42.8 & 39.4 & 77.1 & 58.3 & 33.9 & 37.1 & 43.1 & 73.8 & 51.1 & 70.2 & 32.1 \\
$\lambda_{fmt} = 0.3$ & 47.0 & 51.3 & 36.7 & 33.9 & 73.9 & 55.8 & 29.6 & 34.1 & 36.5 & 75.9 & 43.3 & 64.3 & 29.1 \\
$\lambda_{fmt} = 0.5$ & 45.0 & 48.3 & 34.3 & 32.4 & 74.2 & 53.1 & 26.0 & 27.9 & 39.8 & 75.6 & 41.1 & 58.1 & 29.3 \\
$\lambda_{fmt} = 0.7$ & 48.8 & 53.4 & 37.8 & 35.8 & 74.8 & 56.8 & 32.9 & 35.8 & 39.4 & 75.7 & 46.9 & 66.9 & 29.1 \\

\bottomrule
\end{tabular}
}
\label{tab:reward-abl}
\end{table*}

\begin{table*}[h]
\begin{minipage}{0.99\textwidth}
\begin{AIbox}{System Prompt for model}
\centering
\begin{lstlisting}[basicstyle=\small,]

You are a helpful assistant. When the user asks a question, your response must include two parts: first, the reasoning process enclosed in <think>...</think> tags, then the final answer enclosed in <answer>...</answer> tags. Please provide a clear, concise response within <answer> </answer> tags that directly addresses the question.
\end{lstlisting}
\end{AIbox}
\end{minipage}
\caption{The prompt that used in evaluation.}
\label{tab:eval_prompt}
\end{table*}

\begin{table*}
\begin{minipage}{0.99\textwidth}
\begin{AIbox}{System Prompt for judge model}
\centering
\begin{lstlisting}[basicstyle=\small,]
You are a strict evaluator assessing answer correctness. You must output 1 for fully correct answers and 0 for any other case. You will receive the question, the ground truth answer, and the model prediction.

# Input
Question:
```
{question}
```

Ground Truth Answer:
```
{answer}
```
Model Prediction:
```
{prediction}
```

# Evaluation Rules
- For multiple-choice questions: Score 1 if the predicted answer matches the ground truth answer, it can be directly in option letters or the content of the options.
- For open-ended questions:
  * Score 1 if the prediction matches the answer semantically, it can be in different format.
  * Score 0 for partially correct answers or answers with extra incorrect information, even if the reasoning process is correct.
- Ignore minor differences in formatting, capitalization, or spacing since the model may explain in a different way.
- Treat numerical answers as correct if they match within reasonable precision
- For questions requiring units, both value and unit must be correct

# Strict Output format
1 or 0
\end{lstlisting}
\end{AIbox}
\end{minipage}
\caption{The prompt that used in evaluation.}
\label{tab:judge_prompt}
\end{table*}

\subsection{Training Details}
We describe the training procedures for our best-performing models in both the SFT and RL stages.

\paragraph{SFT.}
To maximize training throughput and reduce memory consumption, we apply online stream packing with an iterable dataset, removing padding tokens to avoid unnecessary computation. We use a packing length of 61,440 tokens. For each batch, we dynamically compute the input token lengths of incoming samples and fill the batch buffer until it is fully packed. Because packing is performed online, the total number of epochs cannot be predetermined; instead, we train each model until convergence. The full set of experimental hyperparameters is provided in ~\cref{tab:train-detail}.

\paragraph{RL.}
For RL training, we use a global batch size of 128, which strikes a balance between on-policy stability and training speed. Log probabilities are computed using a dynamic batch-size implementation without padding to further reduce memory usage. During generation, we set the maximum number of new tokens to 28,696 and cap the prompt length at 4,096 tokens, resulting a max length at 32792. We use a temperature of 1.0 and keep this configuration fixed across all experiments. Due to the high computational cost of RL training, we run training until the reward saturates. The detailed hyperparameters are listed in ~\cref{tab:train-detail}.

\subsection{Evaluation Details}

We now describe our evaluation setup for multimodal reasoning benchmarks. The SFT and RL models share the same evaluation configuration. We use the system prompt shown in ~\cref{tab:eval_prompt} to ensure the model outputs both the reasoning trace and the final answer in an extractable format. The extracted answer is then validated using a two-stage process: a rule-based validator followed by an LLM-as-judge validator. We first apply the rule-based validator to minimize evaluation cost; if the answer cannot be verified, we fall back to the LLM-as-judge, using the prompt provided in ~\cref{tab:judge_prompt}. For all evaluations, we use a temperature of 0.0 for reproducibility and set the maximum generation length to 49K tokens, except for AIME, where we use a temperature of 1.0. We employ vLLM~\citep{kwon2023efficient} as the serving engine to accelerate inference.

\section{Additional Result and Analysis}
\paragraph{Sampling Scaling Results.}
We present the full evaluation results for different sampling strategies in ~\cref{tab:extra_exp}. As shown in the table, scaling up the sampling strategy improves the average score from 46.7 to 49.2, demonstrating the effectiveness of increasing answer diversity along this axis.

\paragraph{Rollout Analysis.}
During RL training, we record all rollout logs and analyze the proportion of reflection-related words in the model’s outputs. We observe that as the reward increases over training steps, the proportion of reflection words also rises. To illustrate this, we generate a word cloud from the final-step rollout outputs—after removing noise words—and find that reflection cues such as “let,” “wait,” and “think” appear frequently in the model’s responses. This indicates that our RL recipe effectively encourages the model to engage in more explicit reasoning as its capabilities improve. The results are shown in ~\cref{fig:rollout_analysis}.

\paragraph{Reward $\lambda_{fmt}$ Ablation.}
To assess the impact of the $\lambda_{fmt}$ parameter in RL, we conduct additional experiments using four values: 0.1, 0.3, 0.5, and 0.7, under the same experimental setup as the GRPO configuration. The evaluation results are presented in ~\cref{tab:reward-abl}. We observe that a lower format-reward weight, specifically $\lambda_{fmt}=0.1$, consistently yields the best performance. Consequently, we adopt $\lambda_{fmt}=0.1$ in our final configuration.

\section{Examples}

\paragraph{Data examples.} We present several examples in~\cref{tab:case-data-1} and~\cref{tab:case-data-2} from our reasoning dataset to demonstrate the high quality of the answer traces generated by our method. These cases illustrate how the model decomposes complex questions into structured reasoning steps and produces coherent, verifiable conclusions.

\paragraph{Qualitative Results.} We present several qualitative examples in~\cref{tab:case-mmmu-pro-vision-1} and~\cref{tab:case-mmmu-pro-vision-2} to demonstrate the effectiveness and robustness of our model. These examples highlight the model’s ability to accurately interpret complex multimodal inputs, generate coherent reasoning steps, and produce reliable answers across diverse scenarios, showcasing both its generalization capability and practical utility.
\section{Limitation and Future Work}

Our work primarily focuses on a single model family—Qwen2.5-VL-Instruct—and evaluates performance mainly within the image domain. While our approach demonstrates strong gains in multimodal reasoning, it does not extend to other modalities such as video or audio, limiting its applicability to a broader set of real-world tasks. Additionally, although we explore scaling strategies in both SFT and RL stages, we have not yet identified the upper bound of model performance under further scaling, leaving open the question of how far the current recipe can be pushed.

For future work, we aim to extend our methodology to a wider range of domains, including video, audio, and richer temporal-stream modalities. Another promising direction is to explore generation capabilities across multiple modalities simultaneously, enabling more coherent and context-aware multimodal reasoning and content synthesis. Broadening the model family used for evaluation and training will further help validate the generality and robustness of our findings across architectures and modality configurations.

\begin{table*}[htp]
\begin{minipage}{0.99\textwidth}
\begin{AIbox}{ColdStart Data Samples}
\centering
\scalebox{0.80}{
\begin{tabular}{l p{13.5cm}}
&  \includegraphics[width=13.5cm]{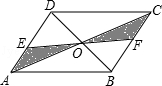} \\
\footnotesize
Question & As shown in the figure, line EF passes through the intersection point O of the diagonals of the parallelogram and intersects AD and BC at points E and F, respectively. If the area of the parallelogram is 12, the sum of the areas of triangles AOE and BOF equals ( )A. 2 B. 3 C. 4 D. Cannot be determined Output the thinking process in <think> </think> and final answer in <answer> </answer> tags. \\
\midrule
\footnotesize
Answer & \begin{CJK}{UTF8}{goth}<think>In a parallelogram, the diagonals bisect each other, so point O is the midpoint of both diagonals AC and BD. Also, since EF passes through O and intersects AD and BC at E and F respectively, we can use symmetry and properties of parallelograms. Because O is the center of the parallelogram, any line through O divides the parallelogram into two regions of equal area. Moreover, triangles AOE and COF are congruent (by ASA or SAS, since AO = OC, ∠AOE = ∠COF, and ∠OAE = ∠OCF due to parallel lines and transversal), and similarly, triangles BOF and DOE are congruent.Therefore, the area of triangle AOE equals the area of triangle COF, and the area of triangle BOF equals the area of triangle DOE.The total area of the parallelogram is 12. The diagonals divide it into 4 triangles of equal area: AOB, BOC, COD, DOA — each of area 3.Now, consider the two shaded triangles: AOE and BOF. Since EF passes through O, and by symmetry, the sum of the areas of AOE and BOF is equal to the area of triangle AOB (which is one of the four equal-area triangles formed by the diagonals).Why? Because:- Area(AOE) + Area(BOF) = Area(AOB)This is because the parts not included in AOE and BOF (namely, EOD and FOC) are congruent to AOE and BOF respectively, so swapping them doesn’t change the total.Alternatively, since the entire parallelogram is divided into 4 triangles of area 3 each, and AOE + BOF together make up exactly one of those triangles (AOB), their combined area is 3.Thus, the answer is 3.</think><answer>B. 3</answer> \end{CJK} \\
\end{tabular}
}
\end{AIbox}
\captionof{table}{An example of the coldstart data sample.}
\vspace{-6mm}
\label{tab:case-data-1}
\end{minipage}
\end{table*}

\begin{table*}[htp]
\begin{minipage}{0.99\textwidth}
\begin{AIbox}{ColdStart Data Samples}
\centering
\scalebox{0.80}{
\begin{tabular}{l p{13.5cm}}
&  \includegraphics[width=13.5cm]{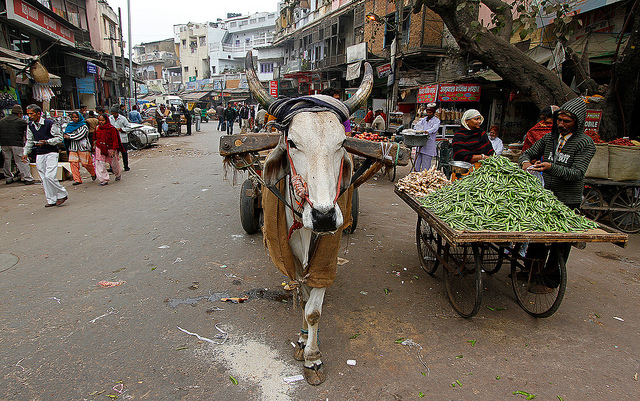} \\
\footnotesize
Question & What is the man doing with the beans?A. cookingB. eating themC. countingD. selling themAnswer with the option's letter from the given choices directly. Output the thinking process in <think> </think> and final answer in <answer> </answer> tags. \\
\midrule
\footnotesize
Answer & \begin{CJK}{UTF8}{goth}<think>The image shows a man sitting beside a cart loaded with green beans in what appears to be a busy street market. He is handling the beans and seems to be interacting with potential customers, which is typical behavior for a vendor. There is no visible cooking equipment, nor is he eating or counting the beans. The context strongly suggests he is selling them.</think><answer>D</answer>\end{CJK} \\
\end{tabular}
}
\end{AIbox}
\captionof{table}{An example of the coldstart data sample.}
\vspace{-6mm}
\label{tab:case-data-2}
\end{minipage}
\end{table*}

\begin{table*}[htp]
\begin{minipage}{0.99\textwidth}
\begin{AIbox}{Qualitative Data Sample}
\centering
\scalebox{0.80}{
\begin{tabular}{l p{13.5cm}}
&  \includegraphics[width=13.5cm]{images/examples/mmmu_pro_0.png} \\
\footnotesize
Question & A. Political instability leading to population decline B. The spread of pathogens across the Silk Road C. Development of new trade routes D. Climate change affecting the Silk Road E. Migrations to areas of Central Asia for resettlement F. Technological advancements in transportationG. Invasions by Mongol tribes H. Large-scale famine due to crop failures I. Economic prosperity and population growth J. Rise of religious conflicts along the Silk Road Answer the question in the image. \\
\midrule
Answer & B \\
\midrule
\footnotesize
Ours & \begin{CJK}{UTF8}{goth}<think>The graph shows population trends for Roman and Han populations over time, with a general decline in population for both over centuries. This decline is most plausibly explained by widespread disease, as the Silk Road facilitated the spread of pathogens across regions, leading to population reductions. While political instability or other factors could contribute, the most direct and historically supported cause for such a widespread population decline across empires is the spread of pathogens.</think><answer>B. The spread of pathogens across the Silk Road</answer>\end{CJK} \\
\midrule
Qwen2.5-VL-7B-Instruct & \begin{CJK}{UTF8}{goth}<think>The graph shows a general decline in population for both the Roman and Han empires over the period from 200 C.E. to 600 C.E. This trend is not consistent with economic prosperity, technological advancements, or large-scale famine, which would typically lead to population growth. The spread of pathogens across the Silk Road could potentially cause a decline in population, but the graph does not provide specific information about the cause of the decline. The rise of religious conflicts along the Silk Road could also lead to population decline, but the graph does not provide specific information about the cause of the decline. The development of new trade routes and the rise of religious conflicts along the Silk Road are not directly related to the population decline shown in the graph. The graph does not provide specific information about the cause of the decline, but it is reasonable to infer that the decline in population could be due to political instability, which could lead to a decline in the overall population.</think><answer>A. Political instability leading to population decline</answer>\end{CJK}
\end{tabular}
}
\end{AIbox}
\vspace{-3mm}
\captionof{table}{An example of our model inference result}
\vspace{-6mm}
\label{tab:case-mmmu-pro-vision-1}
\end{minipage}
\end{table*}

\begin{table*}[htp]
\begin{minipage}{0.99\textwidth}
\begin{AIbox}{Qualitative Data Sample}
\centering
\scalebox{0.80}{
\begin{tabular}{l p{13.5cm}}
&  \includegraphics[width=11.5cm]{images/examples/mmmu_pro_696.png} \\
\footnotesize
Question & A. <image 3> B. <image 2> C. <image 4> D. <image 5> E. <image 1>  Answer the question in the image. \\
\midrule
Answer & D \\
\midrule
\footnotesize
Ours & 
\begin{CJK}{UTF8}{goth}
<think> The relationship between arousal and performance on a challenging task is typically described by the Yerkes-Dodson Law, which states that performance increases with arousal up to a point, after which further increases in arousal lead to a decline in performance. This forms a bell-shaped curve 2014 peak performance at moderate arousal, with performance decreasing as arousal rises too high or falls too low.  Looking at the options: - (A) Inverted V: performance improves with low arousal, then drops 2014 doesn2019t match. - (B) Straight line decreasing: performance always decreases with arousal 2014 too linear and wrong direction. - (C) Straight line increasing: performance always increases with arousal 2014 incorrect. - (D) Bell curve: performance peaks at moderate arousal 2014 matches Yerkes-Dodson Law. - (E) U-shape: performance improves with low arousal, then drops 2014 opposite of what2019s expected.  Thus, (D) best represents the relationship.  The question asks to choose from A-E, and (D) is labeled as image 5 in the options mapping.  So, the correct answer is D. </think><answer>D. <image 5></answer>
\end{CJK}

\\
\midrule
Qwen2.5-VL-7B-Instruct & 
\begin{CJK}{UTF8}{goth}
<think>The relationship between a person's level of arousal and their performance on a challenging task is often described by the Yerkes-Dodson Law, which suggests that performance increases with arousal up to a certain point and then decreases. This relationship is typically represented by a bell-shaped curve, indicating an optimal level of arousal for peak performance.</think><answer>E</answer>
\end{CJK}

\end{tabular}
}
\end{AIbox}
\vspace{-3mm}
\captionof{table}{An example of our model inference result}
\vspace{-6mm}
\label{tab:case-mmmu-pro-vision-2}
\end{minipage}
\end{table*}

%% file: main.bib
@String(CVPR= {IEEE Conf. Comput. Vis. Pattern Recog.})

@String(CVPR  = {CVPR})

@misc{openaio1_cite,
  title={Learning to Reason with LLMs},
  author={OpenAI},
  year={2025},
  howpublished={\url{https://openai.com/index/learning-to-reason-with-llms/}},
}

@article{dsr1_cite,
  title={Deepseek-r1: Incentivizing reasoning capability in llms via reinforcement learning},
  author={Guo, Daya and Yang, Dejian and Zhang, Haowei and Song, Junxiao and Zhang, Ruoyu and Xu, Runxin and Zhu, Qihao and Ma, Shirong and Wang, Peiyi and Bi, Xiao and others},
  journal={arXiv preprint arXiv:2501.12948},
  year={2025}
}

@article{ai2025m2,
  title={M2-reasoning: Empowering mllms with unified general and spatial reasoning},
  author={AI, Inclusion and Wang, Fudong and Liu, Jiajia and Chen, Jingdong and Zhou, Jun and Ji, Kaixiang and Ru, Lixiang and Guo, Qingpei and Zheng, Ruobing and Li, Tianqi and others},
  journal={arXiv preprint arXiv:2507.08306},
  year={2025}
}

@article{wei2025open,
  title={Open vision reasoner: Transferring linguistic cognitive behavior for visual reasoning},
  author={Wei, Yana and Zhao, Liang and Sun, Jianjian and Lin, Kangheng and Yin, Jisheng and Hu, Jingcheng and Zhang, Yinmin and Yu, En and Lv, Haoran and Weng, Zejia and others},
  journal={arXiv preprint arXiv:2507.05255},
  year={2025}
}

@article{an2025llava,
  title={Llava-onevision-1.5: Fully open framework for democratized multimodal training},
  author={An, Xiang and Xie, Yin and Yang, Kaicheng and Zhang, Wenkang and Zhao, Xiuwei and Cheng, Zheng and Wang, Yirui and Xu, Songcen and Chen, Changrui and Wu, Chunsheng and others},
  journal={arXiv preprint arXiv:2509.23661},
  year={2025}
}

@article{meng2025mm,
  title={Mm-eureka: Exploring the frontiers of multimodal reasoning with rule-based reinforcement learning},
  author={Meng, Fanqing and Du, Lingxiao and Liu, Zongkai and Zhou, Zhixiang and Lu, Quanfeng and Fu, Daocheng and Han, Tiancheng and Shi, Botian and Wang, Wenhai and He, Junjun and others},
  journal={arXiv preprint arXiv:2503.07365},
  year={2025}
}

@misc{openai_o3_o4mini_2025,
  author       = {OpenAI},
  title        = {Introducing {\tt o3} and {\tt o4‐mini}},
  howpublished = {\url{https://openai.com/index/introducing-o3-and-o4-mini/}},
  year         = {2025},
}

@misc{deng2025openvlthinkercomplexvisionlanguagereasoning,
      title={OpenVLThinker: Complex Vision-Language Reasoning via Iterative SFT-RL Cycles}, 
      author={Yihe Deng and Hritik Bansal and Fan Yin and Nanyun Peng and Wei Wang and Kai-Wei Chang},
      year={2025},
      eprint={2503.17352},
      archivePrefix={arXiv},
      primaryClass={cs.CV},
      url={https://arxiv.org/abs/2503.17352}, 
}

@misc{chen2025sftrlearlyinvestigation,
      title={SFT or RL? An Early Investigation into Training R1-Like Reasoning Large Vision-Language Models}, 
      author={Hardy Chen and Haoqin Tu and Fali Wang and Hui Liu and Xianfeng Tang and Xinya Du and Yuyin Zhou and Cihang Xie},
      year={2025},
      eprint={2504.11468},
      archivePrefix={arXiv},
      primaryClass={cs.CL},
      url={https://arxiv.org/abs/2504.11468}, 
}

@misc{wei2025openvisionreasonertransferring,
      title={Open Vision Reasoner: Transferring Linguistic Cognitive Behavior for Visual Reasoning}, 
      author={Yana Wei and Liang Zhao and Jianjian Sun and Kangheng Lin and Jisheng Yin and Jingcheng Hu and Yinmin Zhang and En Yu and Haoran Lv and Zejia Weng and Jia Wang and Chunrui Han and Yuang Peng and Qi Han and Zheng Ge and Xiangyu Zhang and Daxin Jiang and Vishal M. Patel},
      year={2025},
      eprint={2507.05255},
      archivePrefix={arXiv},
      primaryClass={cs.CV},
      url={https://arxiv.org/abs/2507.05255}, 
}

@misc{wang2025thinklite,
      title={SoTA with Less: MCTS-Guided Sample Selection for Data-Efficient Visual Reasoning Self-Improvement}, 
      author={Xiyao Wang and Zhengyuan Yang and Chao Feng and Hongjin Lu and Linjie Li and Chung-Ching Lin and Kevin Lin and Furong Huang and Lijuan Wang},
      year={2025},
      eprint={2504.07934},
      archivePrefix={arXiv},
      primaryClass={cs.CV},
      url={https://arxiv.org/abs/2504.07934}, 
}

@misc{bai2025qwen25vltechnicalreport,
      title={Qwen2.5-VL Technical Report}, 
      author={Shuai Bai and Keqin Chen and Xuejing Liu and Jialin Wang and Wenbin Ge and Sibo Song and Kai Dang and Peng Wang and Shijie Wang and Jun Tang and Humen Zhong and Yuanzhi Zhu and Mingkun Yang and Zhaohai Li and Jianqiang Wan and Pengfei Wang and Wei Ding and Zheren Fu and Yiheng Xu and Jiabo Ye and Xi Zhang and Tianbao Xie and Zesen Cheng and Hang Zhang and Zhibo Yang and Haiyang Xu and Junyang Lin},
      year={2025},
      eprint={2502.13923},
      archivePrefix={arXiv},
      primaryClass={cs.CV},
      url={https://arxiv.org/abs/2502.13923}, 
}

@misc{guha2025openthoughtsdatarecipesreasoning,
      title={OpenThoughts: Data Recipes for Reasoning Models}, 
      author={Etash Guha and Ryan Marten and Sedrick Keh and Negin Raoof and Georgios Smyrnis and Hritik Bansal and Marianna Nezhurina and Jean Mercat and Trung Vu and Zayne Sprague and Ashima Suvarna and Benjamin Feuer and Liangyu Chen and Zaid Khan and Eric Frankel and Sachin Grover and Caroline Choi and Niklas Muennighoff and Shiye Su and Wanjia Zhao and John Yang and Shreyas Pimpalgaonkar and Kartik Sharma and Charlie Cheng-Jie Ji and Yichuan Deng and Sarah Pratt and Vivek Ramanujan and Jon Saad-Falcon and Jeffrey Li and Achal Dave and Alon Albalak and Kushal Arora and Blake Wulfe and Chinmay Hegde and Greg Durrett and Sewoong Oh and Mohit Bansal and Saadia Gabriel and Aditya Grover and Kai-Wei Chang and Vaishaal Shankar and Aaron Gokaslan and Mike A. Merrill and Tatsunori Hashimoto and Yejin Choi and Jenia Jitsev and Reinhard Heckel and Maheswaran Sathiamoorthy and Alexandros G. Dimakis and Ludwig Schmidt},
      year={2025},
      eprint={2506.04178},
      archivePrefix={arXiv},
      primaryClass={cs.LG},
      url={https://arxiv.org/abs/2506.04178}, 
}

@article{hurst2024gpt,
  title={Gpt-4o system card},
  author={Hurst, Aaron and Lerer, Adam and Goucher, Adam P and Perelman, Adam and Ramesh, Aditya and Clark, Aidan and Ostrow, AJ and Welihinda, Akila and Hayes, Alan and Radford, Alec and others},
  journal={arXiv preprint arXiv:2410.21276},
  year={2024}
}

@article{comanici2025gemini,
  title={Gemini 2.5: Pushing the frontier with advanced reasoning, multimodality, long context, and next generation agentic capabilities},
  author={Comanici, Gheorghe and Bieber, Eric and Schaekermann, Mike and Pasupat, Ice and Sachdeva, Noveen and Dhillon, Inderjit and Blistein, Marcel and Ram, Ori and Zhang, Dan and Rosen, Evan and others},
  journal={arXiv preprint arXiv:2507.06261},
  year={2025}
}

@misc{zhang2024mathversedoesmultimodalllm,
      title={MathVerse: Does Your Multi-modal LLM Truly See the Diagrams in Visual Math Problems?}, 
      author={Renrui Zhang and Dongzhi Jiang and Yichi Zhang and Haokun Lin and Ziyu Guo and Pengshuo Qiu and Aojun Zhou and Pan Lu and Kai-Wei Chang and Peng Gao and Hongsheng Li},
      year={2024},
      eprint={2403.14624},
      archivePrefix={arXiv},
      primaryClass={cs.CV},
      url={https://arxiv.org/abs/2403.14624}, 
}

@misc{qiao2024wemathdoeslargemultimodal,
      title={We-Math: Does Your Large Multimodal Model Achieve Human-like Mathematical Reasoning?}, 
      author={Runqi Qiao and Qiuna Tan and Guanting Dong and Minhui Wu and Chong Sun and Xiaoshuai Song and Zhuoma GongQue and Shanglin Lei and Zhe Wei and Miaoxuan Zhang and Runfeng Qiao and Yifan Zhang and Xiao Zong and Yida Xu and Muxi Diao and Zhimin Bao and Chen Li and Honggang Zhang},
      year={2024},
      eprint={2407.01284},
      archivePrefix={arXiv},
      primaryClass={cs.AI},
      url={https://arxiv.org/abs/2407.01284}, 
}

@misc{li2024llavaonevisioneasyvisualtask,
      title={LLaVA-OneVision: Easy Visual Task Transfer}, 
      author={Bo Li and Yuanhan Zhang and Dong Guo and Renrui Zhang and Feng Li and Hao Zhang and Kaichen Zhang and Peiyuan Zhang and Yanwei Li and Ziwei Liu and Chunyuan Li},
      year={2024},
      eprint={2408.03326},
      archivePrefix={arXiv},
      primaryClass={cs.CV},
      url={https://arxiv.org/abs/2408.03326}, 
}

@misc{leng2025mmr1enhancingmultimodalreasoning,
      title={MMR1: Enhancing Multimodal Reasoning with Variance-Aware Sampling and Open Resources}, 
      author={Sicong Leng and Jing Wang and Jiaxi Li and Hao Zhang and Zhiqiang Hu and Boqiang Zhang and Yuming Jiang and Hang Zhang and Xin Li and Lidong Bing and Deli Zhao and Wei Lu and Yu Rong and Aixin Sun and Shijian Lu},
      year={2025},
      eprint={2509.21268},
      archivePrefix={arXiv},
      primaryClass={cs.CV},
      url={https://arxiv.org/abs/2509.21268}, 
}

@misc{huan2025doesmathreasoningimprove,
      title={Does Math Reasoning Improve General LLM Capabilities? Understanding Transferability of LLM Reasoning}, 
      author={Maggie Huan and Yuetai Li and Tuney Zheng and Xiaoyu Xu and Seungone Kim and Minxin Du and Radha Poovendran and Graham Neubig and Xiang Yue},
      year={2025},
      eprint={2507.00432},
      archivePrefix={arXiv},
      primaryClass={cs.AI},
      url={https://arxiv.org/abs/2507.00432}, 
}

@misc{wang2025vlrethinkerincentivizingselfreflectionvisionlanguage,
      title={VL-Rethinker: Incentivizing Self-Reflection of Vision-Language Models with Reinforcement Learning}, 
      author={Haozhe Wang and Chao Qu and Zuming Huang and Wei Chu and Fangzhen Lin and Wenhu Chen},
      year={2025},
      eprint={2504.08837},
      archivePrefix={arXiv},
      primaryClass={cs.LG},
      url={https://arxiv.org/abs/2504.08837}, 
}

@misc{qiao2025wemath20versatilemathbook,
      title={We-Math 2.0: A Versatile MathBook System for Incentivizing Visual Mathematical Reasoning}, 
      author={Runqi Qiao and Qiuna Tan and Peiqing Yang and Yanzi Wang and Xiaowan Wang and Enhui Wan and Sitong Zhou and Guanting Dong and Yuchen Zeng and Yida Xu and Jie Wang and Chong Sun and Chen Li and Honggang Zhang},
      year={2025},
      eprint={2508.10433},
      archivePrefix={arXiv},
      primaryClass={cs.AI},
      url={https://arxiv.org/abs/2508.10433}, 
}

@inproceedings{DBLP:conf/cvpr/KembhaviSSCFH17,
  author       = {Aniruddha Kembhavi and
                  Min Joon Seo and
                  Dustin Schwenk and
                  Jonghyun Choi and
                  Ali Farhadi and
                  Hannaneh Hajishirzi},
  title        = {Are You Smarter Than a Sixth Grader? Textbook Question Answering for
                  Multimodal Machine Comprehension},
  booktitle    = {2017 {IEEE} Conference on Computer Vision and Pattern Recognition,
                  {CVPR} 2017, Honolulu, HI, USA, July 21-26, 2017},
  pages        = {5376--5384},
  publisher    = {{IEEE} Computer Society},
  year         = {2017},
  url          = {https://doi.org/10.1109/CVPR.2017.571},
  doi          = {10.1109/CVPR.2017.571},
  bibsource    = {dblp computer science bibliography, https://dblp.org}
}

@misc{ghosal2024algopuzzlevqa,
      title={Are Language Models Puzzle Prodigies? Algorithmic Puzzles Unveil Serious Challenges in Multimodal Reasoning}, 
      author={Deepanway Ghosal and Vernon Toh Yan Han and Chia Yew Ken and Soujanya Poria},
      year={2024},
      eprint={2403.03864},
      archivePrefix={arXiv},
      primaryClass={cs.CV},
      url={https://arxiv.org/abs/2403.03864}, 
}

@misc{chia2024puzzlevqadiagnosingmultimodalreasoning,
      title={PuzzleVQA: Diagnosing Multimodal Reasoning Challenges of Language Models with Abstract Visual Patterns}, 
      author={Yew Ken Chia and Vernon Toh Yan Han and Deepanway Ghosal and Lidong Bing and Soujanya Poria},
      year={2024},
      eprint={2403.13315},
      archivePrefix={arXiv},
      primaryClass={cs.CV},
      url={https://arxiv.org/abs/2403.13315}, 
}

@misc{openr1,
    title = {Open R1: A fully open reproduction of DeepSeek-R1},
    url = {https://github.com/huggingface/open-r1},
    author = {{Hugging Face}},
    month = {January},
    year = {2025}
}

@misc{zhu2025internvl3exploringadvancedtraining,
      title={InternVL3: Exploring Advanced Training and Test-Time Recipes for Open-Source Multimodal Models}, 
      author={Jinguo Zhu and Weiyun Wang and Zhe Chen and Zhaoyang Liu and Shenglong Ye and Lixin Gu and Hao Tian and Yuchen Duan and Weijie Su and Jie Shao and Zhangwei Gao and Erfei Cui and Xuehui Wang and Yue Cao and Yangzhou Liu and Xingguang Wei and Hongjie Zhang and Haomin Wang and Weiye Xu and Hao Li and Jiahao Wang and Nianchen Deng and Songze Li and Yinan He and Tan Jiang and Jiapeng Luo and Yi Wang and Conghui He and Botian Shi and Xingcheng Zhang and Wenqi Shao and Junjun He and Yingtong Xiong and Wenwen Qu and Peng Sun and Penglong Jiao and Han Lv and Lijun Wu and Kaipeng Zhang and Huipeng Deng and Jiaye Ge and Kai Chen and Limin Wang and Min Dou and Lewei Lu and Xizhou Zhu and Tong Lu and Dahua Lin and Yu Qiao and Jifeng Dai and Wenhai Wang},
      year={2025},
      eprint={2504.10479},
      archivePrefix={arXiv},
      primaryClass={cs.CV},
      url={https://arxiv.org/abs/2504.10479}, 
}

@misc{shao2024deepseekmathpushinglimitsmathematical,
      title={DeepSeekMath: Pushing the Limits of Mathematical Reasoning in Open Language Models}, 
      author={Zhihong Shao and Peiyi Wang and Qihao Zhu and Runxin Xu and Junxiao Song and Xiao Bi and Haowei Zhang and Mingchuan Zhang and Y. K. Li and Y. Wu and Daya Guo},
      year={2024},
      eprint={2402.03300},
      archivePrefix={arXiv},
      primaryClass={cs.CL},
      url={https://arxiv.org/abs/2402.03300}, 
}

@misc{yu2025dapoopensourcellmreinforcement,
      title={DAPO: An Open-Source LLM Reinforcement Learning System at Scale}, 
      author={Qiying Yu and Zheng Zhang and Ruofei Zhu and Yufeng Yuan and Xiaochen Zuo and Yu Yue and Weinan Dai and Tiantian Fan and Gaohong Liu and Lingjun Liu and Xin Liu and Haibin Lin and Zhiqi Lin and Bole Ma and Guangming Sheng and Yuxuan Tong and Chi Zhang and Mofan Zhang and Wang Zhang and Hang Zhu and Jinhua Zhu and Jiaze Chen and Jiangjie Chen and Chengyi Wang and Hongli Yu and Yuxuan Song and Xiangpeng Wei and Hao Zhou and Jingjing Liu and Wei-Ying Ma and Ya-Qin Zhang and Lin Yan and Mu Qiao and Yonghui Wu and Mingxuan Wang},
      year={2025},
      eprint={2503.14476},
      archivePrefix={arXiv},
      primaryClass={cs.LG},
      url={https://arxiv.org/abs/2503.14476}, 
}

@misc{zheng2025groupsequencepolicyoptimization,
      title={Group Sequence Policy Optimization}, 
      author={Chujie Zheng and Shixuan Liu and Mingze Li and Xiong-Hui Chen and Bowen Yu and Chang Gao and Kai Dang and Yuqiong Liu and Rui Men and An Yang and Jingren Zhou and Junyang Lin},
      year={2025},
      eprint={2507.18071},
      archivePrefix={arXiv},
      primaryClass={cs.LG},
      url={https://arxiv.org/abs/2507.18071}, 
}

@misc{yue2024mmmumassivemultidisciplinemultimodal,
      title={MMMU: A Massive Multi-discipline Multimodal Understanding and Reasoning Benchmark for Expert AGI}, 
      author={Xiang Yue and Yuansheng Ni and Kai Zhang and Tianyu Zheng and Ruoqi Liu and Ge Zhang and Samuel Stevens and Dongfu Jiang and Weiming Ren and Yuxuan Sun and Cong Wei and Botao Yu and Ruibin Yuan and Renliang Sun and Ming Yin and Boyuan Zheng and Zhenzhu Yang and Yibo Liu and Wenhao Huang and Huan Sun and Yu Su and Wenhu Chen},
      year={2024},
      eprint={2311.16502},
      archivePrefix={arXiv},
      primaryClass={cs.CL},
      url={https://arxiv.org/abs/2311.16502}, 
}

@software{lmms_engine2025,
  title={LMMs Engine: A simple, unified multimodal framework for pretraining and finetuning.},
  author={LMMs-Lab},
  year={2025},
  url={https://github.com/LMMs-Lab/lmms-engine}
}

@misc{zhang2024lmmsevalrealitycheckevaluation,
      title={LMMs-Eval: Reality Check on the Evaluation of Large Multimodal Models}, 
      author={Kaichen Zhang and Bo Li and Peiyuan Zhang and Fanyi Pu and Joshua Adrian Cahyono and Kairui Hu and Shuai Liu and Yuanhan Zhang and Jingkang Yang and Chunyuan Li and Ziwei Liu},
      year={2024},
      eprint={2407.12772},
      archivePrefix={arXiv},
      primaryClass={cs.CL},
      url={https://arxiv.org/abs/2407.12772}, 
}

@misc{lmms_eval2024,
    title={LMMs-Eval: Accelerating the Development of Large Multimoal Models},
    url={https://github.com/EvolvingLMMs-Lab/lmms-eval},
    author={Bo Li* and Peiyuan Zhang* and Kaichen Zhang* and Fanyi Pu* and Xinrun Du and Yuhao Dong and Haotian Liu and Yuanhan Zhang and Ge Zhang and Chunyuan Li and Ziwei Liu},
    publisher    = {Zenodo},
    version      = {v0.1.0},
    month={March},
    year={2024}
}

@inproceedings{kwon2023efficient,
  title={Efficient Memory Management for Large Language Model Serving with PagedAttention},
  author={Woosuk Kwon and Zhuohan Li and Siyuan Zhuang and Ying Sheng and Lianmin Zheng and Cody Hao Yu and Joseph E. Gonzalez and Hao Zhang and Ion Stoica},
  booktitle={Proceedings of the ACM SIGOPS 29th Symposium on Operating Systems Principles},
  year={2023}
}

@article{sheng2024hybridflow,
  title   = {HybridFlow: A Flexible and Efficient RLHF Framework},
  author  = {Guangming Sheng and Chi Zhang and Zilingfeng Ye and Xibin Wu and Wang Zhang and Ru Zhang and Yanghua Peng and Haibin Lin and Chuan Wu},
  year    = {2024},
  journal = {arXiv preprint arXiv: 2409.19256}
}

@inproceedings{
hsu2025ligerkernel,
title={Liger-Kernel: Efficient Triton Kernels for {LLM} Training},
author={Pin-Lun Hsu and Yun Dai and Vignesh Kothapalli and Qingquan Song and Shao Tang and Siyu Zhu and Steven Shimizu and Shivam Sahni and Haowen Ning and Yanning Chen and Zhipeng Wang},
booktitle={Championing Open-source DEvelopment in ML Workshop @ ICML25},
year={2025},
url={https://openreview.net/forum?id=36SjAIT42G}
}

@misc{xu2025llavacotletvisionlanguage,
      title={LLaVA-CoT: Let Vision Language Models Reason Step-by-Step}, 
      author={Guowei Xu and Peng Jin and Ziang Wu and Hao Li and Yibing Song and Lichao Sun and Li Yuan},
      year={2025},
      eprint={2411.10440},
      archivePrefix={arXiv},
      primaryClass={cs.CV},
      url={https://arxiv.org/abs/2411.10440}, 
}

@misc{yang2025r1onevisionadvancinggeneralizedmultimodal,
      title={R1-Onevision: Advancing Generalized Multimodal Reasoning through Cross-Modal Formalization}, 
      author={Yi Yang and Xiaoxuan He and Hongkun Pan and Xiyan Jiang and Yan Deng and Xingtao Yang and Haoyu Lu and Dacheng Yin and Fengyun Rao and Minfeng Zhu and Bo Zhang and Wei Chen},
      year={2025},
      eprint={2503.10615},
      archivePrefix={arXiv},
      primaryClass={cs.CV},
      url={https://arxiv.org/abs/2503.10615}, 
}

@misc{kimiteam2025kimik15scalingreinforcement,
      title={Kimi k1.5: Scaling Reinforcement Learning with LLMs}, 
      author={Kimi Team and Angang Du and Bofei Gao and Bowei Xing and Changjiu Jiang and Cheng Chen and Cheng Li and Chenjun Xiao and Chenzhuang Du and Chonghua Liao and Chuning Tang and Congcong Wang and Dehao Zhang and Enming Yuan and Enzhe Lu and Fengxiang Tang and Flood Sung and Guangda Wei and Guokun Lai and Haiqing Guo and Han Zhu and Hao Ding and Hao Hu and Hao Yang and Hao Zhang and Haotian Yao and Haotian Zhao and Haoyu Lu and Haoze Li and Haozhen Yu and Hongcheng Gao and Huabin Zheng and Huan Yuan and Jia Chen and Jianhang Guo and Jianlin Su and Jianzhou Wang and Jie Zhao and Jin Zhang and Jingyuan Liu and Junjie Yan and Junyan Wu and Lidong Shi and Ling Ye and Longhui Yu and Mengnan Dong and Neo Zhang and Ningchen Ma and Qiwei Pan and Qucheng Gong and Shaowei Liu and Shengling Ma and Shupeng Wei and Sihan Cao and Siying Huang and Tao Jiang and Weihao Gao and Weimin Xiong and Weiran He and Weixiao Huang and Weixin Xu and Wenhao Wu and Wenyang He and Xianghui Wei and Xianqing Jia and Xingzhe Wu and Xinran Xu and Xinxing Zu and Xinyu Zhou and Xuehai Pan and Y. Charles and Yang Li and Yangyang Hu and Yangyang Liu and Yanru Chen and Yejie Wang and Yibo Liu and Yidao Qin and Yifeng Liu and Ying Yang and Yiping Bao and Yulun Du and Yuxin Wu and Yuzhi Wang and Zaida Zhou and Zhaoji Wang and Zhaowei Li and Zhen Zhu and Zheng Zhang and Zhexu Wang and Zhilin Yang and Zhiqi Huang and Zihao Huang and Ziyao Xu and Zonghan Yang and Zongyu Lin},
      year={2025},
      eprint={2501.12599},
      archivePrefix={arXiv},
      primaryClass={cs.AI},
      url={https://arxiv.org/abs/2501.12599}, 
}

@misc{ye2025limoreasoning,
      title={LIMO: Less is More for Reasoning}, 
      author={Yixin Ye and Zhen Huang and Yang Xiao and Ethan Chern and Shijie Xia and Pengfei Liu},
      year={2025},
      eprint={2502.03387},
      archivePrefix={arXiv},
      primaryClass={cs.CL},
      url={https://arxiv.org/abs/2502.03387}, 
}

@misc{li2025miromindm1opensourceadvancementmathematical,
      title={MiroMind-M1: An Open-Source Advancement in Mathematical Reasoning via Context-Aware Multi-Stage Policy Optimization}, 
      author={Xingxuan Li and Yao Xiao and Dianwen Ng and Hai Ye and Yue Deng and Xiang Lin and Bin Wang and Zhanfeng Mo and Chong Zhang and Yueyi Zhang and Zonglin Yang and Ruilin Li and Lei Lei and Shihao Xu and Han Zhao and Weiling Chen and Feng Ji and Lidong Bing},
      year={2025},
      eprint={2507.14683},
      archivePrefix={arXiv},
      primaryClass={cs.CL},
      url={https://arxiv.org/abs/2507.14683}, 
}

@misc{coreteam2025mimovltechnicalreport,
      title={MiMo-VL Technical Report}, 
      author={Core Team and Zihao Yue and Zhenru Lin and Yifan Song and Weikun Wang and Shuhuai Ren and Shuhao Gu and Shicheng Li and Peidian Li and Liang Zhao and Lei Li and Kainan Bao and Hao Tian and Hailin Zhang and Gang Wang and Dawei Zhu and Cici and Chenhong He and Bowen Ye and Bowen Shen and Zihan Zhang and Zihan Jiang and Zhixian Zheng and Zhichao Song and Zhenbo Luo and Yue Yu and Yudong Wang and Yuanyuan Tian and Yu Tu and Yihan Yan and Yi Huang and Xu Wang and Xinzhe Xu and Xingchen Song and Xing Zhang and Xing Yong and Xin Zhang and Xiangwei Deng and Wenyu Yang and Wenhan Ma and Weiwei Lv and Weiji Zhuang and Wei Liu and Sirui Deng and Shuo Liu and Shimao Chen and Shihua Yu and Shaohui Liu and Shande Wang and Rui Ma and Qiantong Wang and Peng Wang and Nuo Chen and Menghang Zhu and Kangyang Zhou and Kang Zhou and Kai Fang and Jun Shi and Jinhao Dong and Jiebao Xiao and Jiaming Xu and Huaqiu Liu and Hongshen Xu and Heng Qu and Haochen Zhao and Hanglong Lv and Guoan Wang and Duo Zhang and Dong Zhang and Di Zhang and Chong Ma and Chang Liu and Can Cai and Bingquan Xia},
      year={2025},
      eprint={2506.03569},
      archivePrefix={arXiv},
      primaryClass={cs.CL},
      url={https://arxiv.org/abs/2506.03569}, 
}

@misc{lu2024mathvistaevaluatingmathematicalreasoning,
      title={MathVista: Evaluating Mathematical Reasoning of Foundation Models in Visual Contexts}, 
      author={Pan Lu and Hritik Bansal and Tony Xia and Jiacheng Liu and Chunyuan Li and Hannaneh Hajishirzi and Hao Cheng and Kai-Wei Chang and Michel Galley and Jianfeng Gao},
      year={2024},
      eprint={2310.02255},
      archivePrefix={arXiv},
      primaryClass={cs.CV},
      url={https://arxiv.org/abs/2310.02255}, 
}

@misc{xiao2024logicvistamultimodalllmlogical,
      title={LogicVista: Multimodal LLM Logical Reasoning Benchmark in Visual Contexts}, 
      author={Yijia Xiao and Edward Sun and Tianyu Liu and Wei Wang},
      year={2024},
      eprint={2407.04973},
      archivePrefix={arXiv},
      primaryClass={cs.AI},
      url={https://arxiv.org/abs/2407.04973}, 
}

@misc{zou2025dynamathdynamicvisualbenchmark,
      title={DynaMath: A Dynamic Visual Benchmark for Evaluating Mathematical Reasoning Robustness of Vision Language Models}, 
      author={Chengke Zou and Xingang Guo and Rui Yang and Junyu Zhang and Bin Hu and Huan Zhang},
      year={2025},
      eprint={2411.00836},
      archivePrefix={arXiv},
      primaryClass={cs.CV},
      url={https://arxiv.org/abs/2411.00836}, 
}

@misc{yue2025mmmuprorobustmultidisciplinemultimodal,
      title={MMMU-Pro: A More Robust Multi-discipline Multimodal Understanding Benchmark}, 
      author={Xiang Yue and Tianyu Zheng and Yuansheng Ni and Yubo Wang and Kai Zhang and Shengbang Tong and Yuxuan Sun and Botao Yu and Ge Zhang and Huan Sun and Yu Su and Wenhu Chen and Graham Neubig},
      year={2025},
      eprint={2409.02813},
      archivePrefix={arXiv},
      primaryClass={cs.CL},
      url={https://arxiv.org/abs/2409.02813}, 
}

@misc{wang2024charxivchartinggapsrealistic,
      title={CharXiv: Charting Gaps in Realistic Chart Understanding in Multimodal LLMs}, 
      author={Zirui Wang and Mengzhou Xia and Luxi He and Howard Chen and Yitao Liu and Richard Zhu and Kaiqu Liang and Xindi Wu and Haotian Liu and Sadhika Malladi and Alexis Chevalier and Sanjeev Arora and Danqi Chen},
      year={2024},
      eprint={2406.18521},
      archivePrefix={arXiv},
      primaryClass={cs.CL},
      url={https://arxiv.org/abs/2406.18521}, 
}

@misc{qwen3vl2025,
  author       = {{Qwen Team}},
  title        = {Qwen3‑VL: Sharper Vision, Deeper Thought, Broader Action},
  howpublished = {\url{https://qwen.ai/blog?id=99f0335c4ad9ff6153e517418d48535ab6d8afef&from=research.latest-advancements-list}},
  year         = {2025},
  note         = {Accessed: 2025‑11‑14}
}

@misc{li2025zebracotdatasetinterleavedvision,
      title={Zebra-CoT: A Dataset for Interleaved Vision Language Reasoning}, 
      author={Ang Li and Charles Wang and Deqing Fu and Kaiyu Yue and Zikui Cai and Wang Bill Zhu and Ollie Liu and Peng Guo and Willie Neiswanger and Furong Huang and Tom Goldstein and Micah Goldblum},
      year={2025},
      eprint={2507.16746},
      archivePrefix={arXiv},
      primaryClass={cs.CV},
      url={https://arxiv.org/abs/2507.16746}, 
}

@misc{yuan2025vlcogitoprogressivecurriculumreinforcement,
      title={VL-Cogito: Progressive Curriculum Reinforcement Learning for Advanced Multimodal Reasoning}, 
      author={Ruifeng Yuan and Chenghao Xiao and Sicong Leng and Jianyu Wang and Long Li and Weiwen Xu and Hou Pong Chan and Deli Zhao and Tingyang Xu and Zhongyu Wei and Hao Zhang and Yu Rong},
      year={2025},
      eprint={2507.22607},
      archivePrefix={arXiv},
      primaryClass={cs.CV},
      url={https://arxiv.org/abs/2507.22607}, 
}

@article{yang2025longvt,
    title={LongVT: Incentivizing "Thinking with Long Videos" via Native Tool Calling},
    author={Yang, Zuhao and Wang, Sudong and Zhang, Kaichen and Wu, Keming and Leng, Sicong and Zhang, Yifan and Li, Bo and Qin, Chengwei and Lu, Shijian and Li, Xingxuan and Bing, Lidong},
    journal={arXiv preprint arXiv:2511.20785},
    year={2025}
}

@misc{zhang2023videollamainstructiontunedaudiovisuallanguage,
      title={Video-LLaMA: An Instruction-tuned Audio-Visual Language Model for Video Understanding}, 
      author={Hang Zhang and Xin Li and Lidong Bing},
      year={2023},
      eprint={2306.02858},
      archivePrefix={arXiv},
      primaryClass={cs.CL},
      url={https://arxiv.org/abs/2306.02858}, 
}

@misc{cheng2024videollama2advancingspatialtemporal,
      title={VideoLLaMA 2: Advancing Spatial-Temporal Modeling and Audio Understanding in Video-LLMs}, 
      author={Zesen Cheng and Sicong Leng and Hang Zhang and Yifei Xin and Xin Li and Guanzheng Chen and Yongxin Zhu and Wenqi Zhang and Ziyang Luo and Deli Zhao and Lidong Bing},
      year={2024},
      eprint={2406.07476},
      archivePrefix={arXiv},
      primaryClass={cs.CV},
      url={https://arxiv.org/abs/2406.07476}, 
}

@misc{zhang2025videollama3frontiermultimodal,
      title={VideoLLaMA 3: Frontier Multimodal Foundation Models for Image and Video Understanding}, 
      author={Boqiang Zhang and Kehan Li and Zesen Cheng and Zhiqiang Hu and Yuqian Yuan and Guanzheng Chen and Sicong Leng and Yuming Jiang and Hang Zhang and Xin Li and Peng Jin and Wenqi Zhang and Fan Wang and Lidong Bing and Deli Zhao},
      year={2025},
      eprint={2501.13106},
      archivePrefix={arXiv},
      primaryClass={cs.CV},
      url={https://arxiv.org/abs/2501.13106}, 
}
